\documentclass{article}

\usepackage{arxiv}
\usepackage[utf8]{inputenc} % allow utf-8 input
\usepackage[T1]{fontenc}    % use 8-bit T1 fonts
\usepackage[colorlinks=true, citecolor=blue,urlcolor=blue]{hyperref}       % hyperlinks
\usepackage{url}            % simple URL typesetting
\usepackage{booktabs}       % professional-quality tables
\usepackage{amsfonts}       % blackboard math symbols
\usepackage{nicefrac}       % compact symbols for 1/2, etc.
\usepackage{microtype}      % microtypography
\usepackage{graphicx}
\usepackage{fontawesome}
\usepackage[font=footnotesize,labelfont=bf]{caption}   
\newcommand{\veclatin}[1]{\bm{#1}} 
\newcommand{\matlatin}[1]{\bm{#1}}
\newcommand{\vecgreek}[1]{\pmb{#1}}

\usepackage{amsmath, amsthm, amssymb}
\usepackage{mathtools}
\usepackage{calc}
\usepackage[flushleft]{threeparttable}
\DeclareMathOperator*{\argmax}{argmax}
\usepackage{bm} 
\usepackage{algorithm}% http://ctan.org/pkg/algorithm
\usepackage{algpseudocode}% http://ctan.org/pkg/algorithmicx
\usepackage{etoolbox}
\usepackage{tikz}
\usetikzlibrary{tikzmark}
\usetikzlibrary{calc}

\errorcontextlines\maxdimen

\newcommand{\ALGtikzmarkcolor}{black}% customise this, if you want
\newcommand{\ALGtikzmarkextraindent}{4pt}% customise this, if you want
\newcommand{\ALGtikzmarkverticaloffsetstart}{-.5ex}% customise this, if you want
\newcommand{\ALGtikzmarkverticaloffsetend}{-.5ex}% customise this, if you want
\makeatletter
\newcounter{ALG@tikzmark@tempcnta}

\newcommand\ALG@tikzmark@start{%
	\global\let\ALG@tikzmark@last\ALG@tikzmark@starttext%
	\expandafter\edef\csname ALG@tikzmark@\theALG@nested\endcsname{\theALG@tikzmark@tempcnta}%
	\tikzmark{ALG@tikzmark@start@\csname ALG@tikzmark@\theALG@nested\endcsname}%
	\addtocounter{ALG@tikzmark@tempcnta}{1}%
}

\def\ALG@tikzmark@starttext{start}
\newcommand\ALG@tikzmark@end{%
	\ifx\ALG@tikzmark@last\ALG@tikzmark@starttext
	\else
	\tikzmark{ALG@tikzmark@end@\csname ALG@tikzmark@\theALG@nested\endcsname}%
	\tikz[overlay,remember picture] \draw[\ALGtikzmarkcolor] let \p{S}=($(pic cs:ALG@tikzmark@start@\csname ALG@tikzmark@\theALG@nested\endcsname)+(\ALGtikzmarkextraindent,\ALGtikzmarkverticaloffsetstart)$), \p{E}=($(pic cs:ALG@tikzmark@end@\csname ALG@tikzmark@\theALG@nested\endcsname)+(\ALGtikzmarkextraindent,\ALGtikzmarkverticaloffsetend)$) in (\x{S},\y{S})--(\x{S},\y{E});%
	\fi
	\gdef\ALG@tikzmark@last{end}%
}

\apptocmd{\ALG@beginblock}{\ALG@tikzmark@start}{}{\errmessage{failed to patch}}
\pretocmd{\ALG@endblock}{\ALG@tikzmark@end}{}{\errmessage{failed to patch}}
\makeatother
\usepackage{setspace}
\let\Algorithm\algorithm
\renewcommand\algorithm[1][]{\Algorithm[#1]\setstretch{1.4}}

\algdef{SE}[SUBALG]{Indent}{EndIndent}{}{\algorithmicend\ }%
\algtext*{Indent}
\algtext*{EndIndent}

\usepackage[authoryear, round]{natbib}
\title{\textbf{XGBoostLSS} \\ \vspace{0.5em} {\large An extension of XGBoost to probabilistic forecasting}} 			

\author{Alexander März$^{*}$\thanks{$^{*}$Address for correspondence: \texttt{alex.maerz@gmx.net.}
	}
	\vspace{-1em}
}

\begin{document} 
	\maketitle
	
	\vspace{2em} 
	
	\begin{abstract}
		We propose a new framework of XGBoost that predicts the entire conditional distribution of a univariate response variable. In particular, \texttt{XGBoostLSS} models all moments of a parametric distribution (i.e., mean, location, scale and shape [LSS]) instead of the conditional mean only. Choosing from a wide range of continuous, discrete and mixed discrete-continuous distribution, modelling and predicting the entire conditional distribution greatly enhances the flexibility of XGBoost, as it allows to gain additional insight into the data generating process, as well as to create probabilistic forecasts from which prediction intervals and quantiles of interest can be derived. We present both a simulation study and real world examples that demonstrate the virtues of our approach. 
	\end{abstract}
	
	\keywords{Bayesian Optimization \and Distributional Modelling \and Expectile Regression \and GAMLSS \and Probabilistic Forecast \and Uncertainty Quantification \and XGBoost}
	
	\vspace{2em}

	\section{Introduction} \label{sec:introduction}
	
	\begin{quote} 
		\it{The ultimate goal of regression analysis is to obtain information about the \textbf{[entire] conditional distribution} of a response given a set of explanatory variables}.\footnote{Emphasize added.}\citep{Hothorn.2014}
	\end{quote}
	
	We couldn't agree more. Yet, many regression models focus on the conditional mean $\mathbb{E}(Y|\mathbf{X} = \mathbf{x})$ only, implicitly treating higher moments of the conditional distribution $F_{Y}(y|\mathbf{x})$ as fixed nuisance parameters.\footnote{We follow \citet{Hothorn.2018} and denote $\mathbb{P}(Y \leq y | \mathbf{X} = \mathbf{x}) = F_{Y}(y|\mathbf{x})$ the conditional distribution of a potentially continuous, discrete or mixed discrete-continuous response $Y$ given explanatory variables $\mathbf{X} = \mathbf{x}$.} This assumption, however, of constant higher moments not changing as functions of covariates is a stark one and is only valid in situations where the user\footnote{To keep the term as broad as possible, we have chosen to use the word 'user', which in term can imply researcher, analyst or data scientist.} is privileged with dealing with data generated by a symmetric Gaussian distribution with constant variance.\footnote{Note that the Gaussian distribution is fully characterised by its first two moments, i.e., mean and variance.} In real world situations, however, the data generating process is usually less well behaved, exhibiting characteristics such as heteroskedasticity, varying degrees of skewness and/or kurtosis. If the user sticks to his/her assumption of not modelling all characteristics of the data, inference as well as uncertainty assessments, such as confidence and predictions intervals, are at best invalid. In this context, the introduction of Generalised Additive Models for Location Scale and Shape (GAMLSS) by \citet{Rigby.2005} has stimulated a lot of research and culminated in a new branch of statistics that focuses on modelling the entire conditional distribution as functions of covariates.
	
	%Furthermore, let $f(y_{i}|\veclatin{\theta}_{i})$ denote a probability density function that depends on $K$ distributional parameters $\veclatin{\theta} = (\theta_{ik}, \ldots, \theta_{iK})^{\prime}, i = 1, \ldots, n$. 
	
	Consulting the literature on computer science and machine learning, however, shows that the main focus so far has been on prediction accuracy and estimation speed. In fact, even though machine learning approaches (e.g., Random Forest or Gradient Boosting-type algorithms) outperform many statistical approaches when it comes to prediction accuracy, the output/forecast of these models provides information about the conditional mean $\mathbb{E}(Y|\mathbf{X} = \mathbf{x})$ only. As a consequence, this class of models is rather reluctant to reveal other characteristics of the (predicted) distribution and falls short in applications where probabilistic forecasts are required, e.g., for assessing prediction uncertainty in form of prediction intervals. This is consistent with the assertion made in \citet{Breiman.2001}, who distinguishes two opposing cultures of statistical modelling: the first is the so called 'Data Modelling Culture' that starts the analysis with assuming a stochastic data model for the algorithm. The parameters of the underlying data generating distribution are estimated and the algorithm is then used for inference and/or prediction. In contrast, there is the so called 'Algorithmic Modelling Culture' that considers the inside of the algorithm complex and unknown, with the aim of estimating a function $f(\mathbf{x})$ to predict the response $Y$. 
	
	While the approaches discussed in \citet{Breiman.2001} are an admissible partitioning of the space of how to analyse and model data, more recent advances have gradually made this distinction less clear-cut (see Section \ref{sec:research} for an overview). In fact, the current research trend in both statistics and machine learning gravitates towards bringing both disciplines closer together. In an era of increasing necessity that the output of prediction models needs to be turned into explainable and reliable insights, this is an exceedingly promising and encouraging development, as both disciplines need and should mutually enrich each other. This paper contributes to further closing the gap between the two cultures by extending statistical boosting to a machine learning approach that accounts for for all distributional properties of the data. In particular, we present an extension of XGBoost introduced by \citet{Chen.2016} which has gained much popularity and attention over the last years and has arguably become among the most widely used tools in practical data science. We term our model \texttt{XGBoostLSS}, as it combines the accuracy and speed of XGBoost with the flexibility and interpretability of GAMLSS that allow for the estimation and prediction of the entire conditional distribution $F_{Y}(y|\mathbf{x})$. \texttt{XGBoostLSS} allows the user to choose from a wide range of continuous, discrete and mixed discrete-continuous distributions to better adapt to the data at hand, as well as to provide predictive distributions, from which prediction intervals and quantiles can be derived. Furthermore, all XGBoost additions, such as partial dependent plots, parallel model training, both CPU and GPU, as well as distributed computing using, e.g., Spark and Dask, fast histogram model training or the recently added SHAP (SHapley Additive exPlanations) approach of \citet{Lundberg.2019} that allows to explain the output of any machine learning model, are still applicable, with the additional advantage that insights can be provided for all distributional parameters. As such, \texttt{XGBoostLSS} is intended to weaken the separation between the 'Data Modelling Culture' and 'Algorithmic Modelling Culture', so that models designed mainly for prediction can also be used to describe and explain the underlying data generating process of the response of interest.
	
	The remainder of this paper is organised as follows: Section \ref{sec:gamlss} introduces the reader to distributional modelling and Section \ref{sec:research} presents an overview of related research branches. In Section \ref{sec:xgboostlss}, we formally introduce \texttt{XGBoostLSS}, while Section \ref{sec:applications} presents both a simulation study and real world examples that provide a walk-through of the functionality of our model. Section \ref{sec:implementation} gives an overview of available software implementations and Section \ref{sec:conclusion} concludes.

	\section{Distributional Modelling} \label{sec:gamlss}
	
	\begin{quote} 
		\it{There is indeed more to life than mean and variance. A good point at which to start is by replacing them by location and scale and noting that one reason for the stress on mean and variance is the implicit assumption of Gaussianity. Once the assumption of Gaussianity is dropped, attention shifts to estimating [all of] the parameter in a distribution.} \citep{Harvey.2013}
	\end{quote}
	
	According to \citet{Stasinopoulos.2015}, there are two important issues in any statistical model: the appropriate choice of a distribution for the response and explaining how the parameters of the assumed distribution change with the explanatory variables. Generalised Additive Models for Location Scale Shape (GAMLSS) introduced by \citet{Rigby.2005} and extended to a Bayesian framework by \citet{Klein.2015b, Klein.2015c} provide such a framework that allows modelling all distribution parameters as functions of covariates. This section introduces the reader to the general idea of distributional modelling. In order to fully understand their beauty and elegance, we draw the reader's attention to \citet{Rigby.2005, Klein.2015b, Klein.2015c, Stasinopoulos.2017}.
	
	In its original formulation, GAMLSS assume that a univariate response follows a distribution $\mathcal{D}$ that depends on up to four parameters, i.e., $y_{i} \stackrel{ind}{\sim} \mathcal{D}(\mu_{i}, \sigma^{2}_{i}, \nu_{i}, \tau_{i}), i=1,\ldots,n$, where $\mu_{i}$ and $\sigma^{2}_{i}$ are location and scale parameters, respectively, while $\nu_{i}$ and $\tau_{i}$ correspond to shape parameters such as skewness and kurtosis. Hence, the framework allows to model not only the mean (or location) but all parameters as functions of explanatory variables. In contrast to Generalised Linear (GLM) and Generalised Additive Models (GAM), the assumption of the response belonging to an exponential family type of distribution is relaxed in GAMLSS and replaced by a general distribution family, including highly skewed and/or kurtotic continuous,  discrete and mixed discrete distributions. From a frequentist point of view, GAMLSS can be formulated as follows: let $\textbf{y} = (y_{1}, \ldots, y_{n})^{\prime}$ be the vector of a univariate response variable, with $g_{k}(\cdot)$ being a known monotonic link function relating the distribution parameters $\theta_{k} = 1, 2, 3, 4$ to explanatory variables:
	\begin{equation}
	\begin{aligned}
	g_{1}(\vecgreek{\mu}) &= \vecgreek{\eta}_{1} = \textbf{X}_{1}\vecgreek{\beta}_{1} + \sum^{p_{1}}_{j=1}f_{1,j}(\veclatin{z}_{j})  \qquad \qquad 
	g_{2}(\vecgreek{\sigma^{2}}) &= \vecgreek{\eta}_{2} = \textbf{X}_{2}\vecgreek{\beta}_{2} + \sum^{p_{2}}_{j=1}f_{2,j}(\veclatin{z}_{j}) \\
	g_{3}(\vecgreek{\nu}) &= \vecgreek{\eta}_{3} = \textbf{X}_{3}\vecgreek{\beta}_{3} + \sum^{p_{3}}_{j=1}f_{3,j}(\veclatin{z}_{j}) \qquad \qquad 
	g_{4}(\vecgreek{\tau}) &= \vecgreek{\eta}_{4} = \textbf{X}_{4}\vecgreek{\beta}_{4} + \sum^{p_{4}}_{j=1}f_{4,j}(\veclatin{z}_{j}) 
	\end{aligned}
	\end{equation} 
	\noindent where $\vecgreek{\beta}_{k} = (\beta_{1k}, \beta_{2k}, \ldots, \beta_{qk})^{\prime}$ is a parameter vector modelling linear effects or categorical variables, $\textbf{X}_{k}$ is the corresponding design matrix and $f_{k,j}(\veclatin{z}_{j})$ reflect different types of regression effects that model the effect of a continuous covariate $\veclatin{z}_{j}$. The flexibility of the GAMLSS framework comes from its modelling of all distribution parameters of $\mathcal{D}(\vecgreek{\mu}(\veclatin{x}), \vecgreek{\sigma}^{2}(\veclatin{x}), \vecgreek{\nu}(\veclatin{x}), \vecgreek{\tau}(\veclatin{x})) = \mathcal{D}(\veclatin{\theta}(\veclatin{x}))$ and from approximating $f_{j}$ in terms of basis function expansions (see \citealt{Fahrmeir.2011} and \citealt{Fahrmeir.2013} for details):\footnote{Without loss of generality, the index $k$ that indicates the distributional parameter is dropped for notational simplicity.}
	
	\begin{equation}
	f_{j}(\veclatin{z}_{j}) = \sum^{D_{j}}_{d_{j}=1}\gamma_{j,d_{j}}B_{j,d_{j}}(\veclatin{z}_{j}) 
	\end{equation}
	where $B_{j,d_{j}}(\veclatin{z}_{j})$ are basis functions and $\gamma_{j,d_{j}}$ denote the corresponding basis coefficients. Corresponding to each function $f_{j}$, there is a quadratic penalty term attached
	\begin{equation}
	\mbox{pen}(f_{j}) = \lambda_{j} \vecgreek{\gamma}^{\prime}_{j}\matlatin{K}_{j}\vecgreek{\gamma}_{j} \label{eq:pen}  
	\end{equation}
	\noindent that enforces specific properties of the function such as smoothness, where $\vecgreek{\gamma}_{j} = (\gamma_{j,1}, \ldots, \gamma_{j,d_{j}})^{\prime}$ is a vector of basis coefficients, $\matlatin{K}_{j}$ is a penalty matrix and $\lambda_{j}\geq0$ is a smoothing parameter that governs the impact of the penalty. Besides the modelling of each parameter of a wide range of distributions within a regression setting\footnote{GAMLSS currently provide over 80 continuous, discrete and mixed distributions for modelling the response variable}, the GAMLSS framework allows incorporating numerous covariate specifications in the modelling process and comprises several well-known special cases such as Generalized Linear Models \citep{Nelder.1972}, Generalized Additive (Mixed) Models \citep{Hastie.1990, Lin.1999}, Varying Coefficient Models \citep{Hastie.1993} or Geoadditive Models \citep{Kammann.2003}. Even though we follow the naming GAMLSS of \citet{Rigby.2005}, it is not necessarily true that the distribution at hand is characterized by parameters that represent shape parameters, i.e., skewness and kurtosis. Hence, we follow \citep{Klein.2015b} and use the term distributional modelling and GAMLSS interchangeably. Concerning the estimation of GAMLSS, it relies on the availability of first and second order derivatives of the (log)-likelihood function needed for Fisher-scoring type algorithms. As we will see in Section \ref{sec:xgboostlss}, this is very closely related to the estimation of XGBoost, which we will exploit to arrive at \texttt{XGBoostLSS}.
	
	We would like to draw the attention of the reader to an implication that is a consequence of modelling and predicting the entire distribution. Standard regression/supervised models assume the observations to be independent and identically distributed (iid) realizations $y \stackrel{iid}{\sim} \mathcal{D}(\veclatin{\theta})$, where $\veclatin{\theta}$ is a vector of distributional parameters. In contrast, however, distributional modelling implies that the observations are independent, but not necessarily identical realizations $y \stackrel{ind}{\sim} \mathcal{D}(\veclatin{\theta}(\veclatin{x}))$, where all distributional parameters $\veclatin{\theta}(\veclatin{x})$ are related to and allowed to change with covariates. To illustrate the implications of distributional modelling, let us re-visit the concept of stationarity used in time series analysis, with covariates $\veclatin{x}$ including time.\footnote{A nice statement that summarises the concept of stationarity is made by \citet{Albran.1974}: \textit{'I have seen the future and it is very much like the present, only longer}.'} Most forecasting methods are based on the assumption that the time series at hand can be rendered approximately stationary through the use of appropriate transformations, e.g., difference-stationary or trend-stationary. In general, one can distinguish two forms of stationarity. The first, and the weaker one, is covariance stationarity which requires the first moment (i.e., the mean) and auto-covariance to not vary with respect to time. The second, and more strict one, is strong stationarity that can be formulated as follows
	
	\begin{equation}
	F_{Y}(y_{t_{1}}, \ldots, y_{t_{n}}) = F_{Y}(y_{t_{1} + \tau }, \ldots, y_{t_{n} + \tau}), \qquad \forall \mbox{ } n, t_{1}, \ldots, t_{n}, \tau 
	\end{equation}
	
	\noindent where $F_{Y}(\cdot)$ is the joint cumulative distribution function of $\left\lbrace y_{t}\right\rbrace $ at times $t$. Given that $F_{Y}(\cdot)$ does not change with a shift in time of $\tau$, it follows that all parameters of a strictly stationary process are time invariant. However, this is a very restrictive assumption that is likely to be violated in many real world situations. As all distributional parameters are functions of covariates, distributional modelling is able to account for the non-stationarity so that stationarity does not need to serve as default assumption in applied modelling. \footnote{However, we also stress that non-stationarity modelling is an option and does not provide a universal solution that should blindly be applied without any support from the data. For a discussion on non-stationarity modelling in hydrologic flood frequency analyses and climate change modelling see \citet{Villarini.2009}, \citet{Milly.2015} and \citet{Serinaldi.2015}.}
	
	As an additional initiative towards highlighting the insights one can generate with modelling all parameters of a response distribution, 
	we want to emphasize that distributional modelling can make valuable contributions to a recent strand of literature in social science and economics that highlights the importance of analysing conditional heteroskedasticity in addition to the conditional mean only. Contrary to the commonly held view that heteroskedasticity is only relevant when it comes to alleviating adverse effects on statistical inference, we follow the works of \citet{Downs.1979}, \citet{Western.2009}, \citet{Zheng.2013} and consider its analysis to be an important source of revealing additional information that would otherwise go undetected. In general, heteroskedastic regression models are intended to model the conditional variance of the response variable as a function of covariates within a regression setting, instead of treating it as a nuisance only. Extensions of the conventional regression models, termed heteroskedastic regression models (HRM, \citealt{Smyth.1989}), variance function regression models \citep{Western.2009}, double generalized linear models (DGLM, \citealt{Smyth.2001, Smyth.2002}) or double hierarchical generalized linear models (DHGLM, \citealt{Nelder.1991}), have recently been used in sociology and economics not only to detect violations of standard regression assumptions, but also for substantive insight. An example of this includes the excess residual variation in income inequality within certain population subgroups that has been interpreted as reflecting unobserved skills or economic insecurity (see, e.g., \citealt{Western.2009}). As an illustration, consider gender as a categorical covariate that has two groups, male and female. In a standard conditional mean setting, regression coefficients describe differences in group means, e.g., expected differences in monthly salaries between men and women. In addition to analysing these between-group differences, heteroskedastic regression models extend the analysis to within-group differences, i.e., testing heterogeneity within groups, for example within men and within woman, for systematic differences. In other words, covariate effects in conditional mean regression account for deviations of the group sample means from the overall mean of the response (between-group differences), while covariate effects in heteroskedastic regression models explain how the variability of the response around group means changes as a function of covariates within groups (within-group difference) \citep{Zheng.2011}. Consequently, parallel to studying between-group differences within a conditional mean regression setting, the analysis of within-group heterogeneity modelled as conditional heteroskedasticity yields a more complete picture of the response variable \citep{Zheng.2013}. With respect to giving an economic interpretation of within-group differences in the form of heteroskedasticity, the literature on income inequality has offered the interpretation of heteroskedasticity as reflecting the influence of unobserved or hidden heterogeneity in the form of luck \citep{Jencks.1972}, skill, such as intrinsic ability, work effort and school quality \citep{Juhn.1993, Lemieux.2006}, or as measuring income risk and insecurity \citep{Western.2008, Western.2009}.

	\section{Related Research} \label{sec:research}
	
	Reviewing the current literature at the intersect between machine learning/computer science and statistics shows that there has been an incredibly rich stream of ideas that aim at bringing the two disciplines closer together. As this section cannot give an exhaustive overview of all approaches, it focuses on some selected recent advances only, with a particular focus on statistical boosting, as this branch of statistics is most closely related to our approach. 
	
	In fact, statistical boosting evolved out of machine learning and was adapted to estimate classical statistical models \citep{Mayr.2017}. Among a great variety of approaches, probably among the most powerful class of models is component-wise gradient descent boosting of \citep{Breiman.1998, Breiman.1999, Friedman.2000, Friedman.2001} that estimates statistical models via gradient descent, most prominently Generalized Additive Models implemented in $\it{mboost}$ of \citep{Buehlmann.2007, Hothorn.2010, Hofner.2014, Hofner.2015, Hothorn.2018}.\footnote{\citet{Schalk.2018} provide an alternative implementation of component-wise boosting written in \texttt{C++} to obtain high runtime performance. However, a GAMLSS implementation is not yet available.} The approach, however, that is closest to \texttt{XGBoostLSS}, is $\it{gamboostLSS}$ of \citep{Mayr.2012, Hofner.2016, Hofner.2018, Thomas.2018}, that allows to fit GAMLSS via component-wise boosting. In fact, \texttt{XGBoostLSS} and $\it{gamboostLSS}$ are closely related as both of them extend GAMLSS to boosting-type approaches. However, arguably the key difference between \texttt{XGBoostLSS} and $\it{gamboostLSS}$ is that the latter takes a statistical boosting point of view and is designed to estimate classical regression models, while \texttt{XGBoostLSS} originates in pure machine learning and computer science. As such, it is optimized for prediction accuracy and high performance computing, which makes a significant impact when it comes to factorization of use cases, where estimation speed is very often as important as prediction accuracy. In particular, besides its inherent parallelization, the availability of several Spark interfaces of XGBoost that enable training over distributed datasets makes XGBoost in general, and \texttt{XGBoostLSS} in particular suited for handling large datasets. Another advantage of \texttt{XGBoostLSS} is that XGBoost is currently available for several programming languages such as \texttt{R}, \texttt{Python}, \texttt{Julia} and \texttt{Scala}, while $\it{gamboostLSS}$ is implemented in \texttt{R} only. It is important to stress, however, that the fact as such that $\it{gamboostLSS}$ and \texttt{XGBoostLSS} originate from different backgrounds does not make one approach superior to the other. The choice of which approach to use depends, as always, on the purpose and problem at hand. While existing GAMLSS frameworks and implementations are supposed to perform well for small to medium sized data sets, \texttt{XGBoostLSS} plays off its strengths in situations where the user faces data sets that deserve the term big data. The motivation for a distributed and scalable extension of statistical boosting is nicely summarized in the following statement:
	
	%\vspace{-0.4em}
	
	\begin{quote} 
		\it{Regarding future research, a huge challenge for the use of boosting algorithms in biomedical applications arises from the era of big data. Unlike other machine learning methods like random forests, the sequential nature of boosting methods hampers the use of parallelization techniques within the algorithm, which may result in issues with the fitting and tuning of complex models with multidimensional predictors and/or sophisticated base-learners like splines or higher-sized trees. To overcome these problems in classification and univariate regression, \citet{Chen.2016} developed the extremely fast and sophisticated xgboost environment. \textbf{For the more recent extensions discussed in this paper, however, big data solutions for statistical boosting have yet to be developed}.}\footnote{Emphasize added.}\citep{Mayr.2017}
	\end{quote}
	
	%\vspace{-0.4em}
	
	Based on conditional inference trees and forests of \citep{Hothorn.2006, Zeileis.2008, Hothorn.2015}, \citet{Schlosser.2018, Schlosser.2019} recently introduced Distributional Regression Forests that extend GAMLSS using Random Forests. As with statistical boosting, we consider conditional inference trees and forests in general and Distributional Regression Forests in particular belonging to the area of statistical models, as they embed tree-structured regression models into a well defined theory of conditional inference procedures, where significance tests are used for recursive partitioning, which makes estimation slow and not applicable for large data sets. Other recent approaches are Quantile Regression Forests introduced by \citep{Meinshausen.2006, Meinshausen.2017} and Generalised Regression Forest of \citep{Athey.2019, Tibshirani.2018} that use a local nearest neighbour weights approach to estimate different points of the conditional distribution. Bayesian Additive Regression Trees (BART) of \citep{Chipman.2010, McCulloch.2019} are another very interesting strand of literature, as they take a Bayesian view of estimating decision trees and forests. To accommodate for heteroskedastic settings, \citet{Pratola.2018} recently introduced a heteroscedastic version of BART.
	
	Among several other interesting approaches that focus on distributional modelling, we would like to highlight Conditional Transformation Models (CTMs) introduced by \citet{Hothorn.2014}. In a nutshell, CTM model the conditional distribution function $\mathbb{P}(Y \leq y | \mathbf{X} = \mathbf{x}) = F_{Y}(y|\mathbf{x}) = F(h(y|\mathbf{x}))$ of a response $Y$ in terms of a monotone transformation function $h: \mathbb{R} \rightarrow \mathbb{R}$, where $F(\cdot)$ denotes an continuous cumulative distribution function $F \rightarrow \mathbb{R}  [0, 1]$ with corresponding quantile function $Q = F^{-1}(\cdot)$, where the transformation function $h$ is allowed to depend on explanatory variables $\mathbf{x}$. Intuitively, CTMs can be understood as the inverse of a quantile regression model, as they model the conditional distribution function of the responses directly \citep{Hothorn.2018}. Hence, CTMs are able to estimate all quantiles simultaneously in a joint model, which is in contrast to quantile regression were separate models are estimated for different quantiles. In CTMs, the transformation function $h$ is estimated semi-parametrically under rather weak assumptions. Recently, \citep{Hothorn.2019, Hothorn.2018a} and \citep{Hothorn.2019b, Hothorn.2019c} extended CTMs to Transformation Forests and Transformation Boosting Machines, respectively, with \citet{Klein.2019} introducing multivariate conditional transformation models.

	\section{XGBoostLSS} \label{sec:xgboostlss} 
	
	In this section, we introduce \texttt{XGBoostLSS}. As our model is based on XGBoost, we also briefly touch upon its functioning, while referring the interested reader to \citet{Chen.2016} for a more detailed exposition.\footnote{A very accessible introduction to XGBoost can be found in \citet{Nielsen.2016}.} In XGBoost, the estimation at each iteration $t$ is based on minimizing the following regularized objective function
	
	\begin{equation}
	\begin{aligned}
	\mathcal{L}^{(t)} &= \sum^{n}_{i=1} \ell[y_{i}, \hat{y}^{(t)}_{i}]  + \Omega(f_{t}) \\
	&= \sum^{n}_{i=1} \ell[y_{i}, \hat{y}^{(t-1)}_{i} + f_{t}(\mathbf{x}_{i})] + \Omega(f_{t}) \label{eq:xgboost1} 
	\end{aligned}
	\end{equation}
	
	\noindent where $l$ is a differentiable convex loss function that measures the discrepancy between the prediction of the $i$-th instance at the $t$-th iteration $\hat{y}^{(t)}_{i} = \hat{y}^{(t-1)}_{i} + f_{t}(\mathbf{x}_{i})$ and the true value $y_{i}$, while $\Omega(\cdot)$ is a regularization term that penalizes the complexity of the model to avoid over-fitting. A second order approximation of $\ell[\cdot]$ and dropping constant terms allows to re-write Equation \eqref{eq:xgboost1}
	
	\begin{equation}
	\mathcal{\tilde{L}}^{(t)} = \sum^{n}_{i=1} [g_{i}f_{t}(\mathbf{x}_{i}) + \frac{1}{2}h_{i}f^{2}_{t}(\mathbf{x}_{i})] + \Omega(f_{t}) \label{eq:xgboost2} 
	\end{equation}
	
	where $g_{i} = \partial_{\hat{y}^{(t-1)}}\ell[y_{i}, \hat{y}^{(t-1)}_{i}]$ and $h_{i} = \partial^{2}_{\hat{y}^{(t-1)}}\ell[y_{i}, \hat{y}^{(t-1)}_{i}]$ are first and second order derivatives of the loss w.r.t. its second argument evaluated at $[y_{i}, \hat{y}^{(t-1)}_{i}]$. Expanding $\Omega(\cdot)$, we can re-write Equation \eqref{eq:xgboost2} as follows
	
	\begin{equation}
	\mathcal{\tilde{L}}^{(t)} = \sum^{n}_{i=1} [g_{i}f_{t}(\mathbf{x}_{i}) + \frac{1}{2}h_{i}f^{2}_{t}(\mathbf{x}_{i})] + \gamma T + \frac{1}{2} \lambda \sum^{T}_{j=1}w^{2}_{j} \label{eq:xgboost_final} 
	\end{equation}
	
	%\begin{equation}
	%w_{j} = - \frac{G_{j}}{H_{j} + \lambda} 
	%\end{equation}
	
	\noindent where $w_{j}$ are leaf weights, $\gamma$ is a parameter that controls the penalization for the number of terminal nodes $T$ of the trees and $\lambda$ is a $L_{2}$ regularization term on the leaf weights.\footnote{One can further re-write Equation \eqref{eq:xgboost_final} and calculate the optimal weights $w^{*}_{j}$. For more details see \citet{Chen.2016} and 	\href{https://xgboost.readthedocs.io/en/latest/tutorials/model.html}{https://xgboost.readthedocs.io/en/latest/tutorials/model.html}.}
	
	There are several characteristics that set XGBoost apart from other existing boosting approaches. The first is its implicit regularization of the complexity of the trees, that prevents it from over-fitting. More importantly, however, is that XGBoost is based on Newton boosting, also called second order gradient boosting. As we see from Equation \eqref{eq:xgboost_final}, the loss function $\ell[\cdot]$ is approximated by a second order Taylor expansion, where in each iteration $t$, the first and second order partial derivatives of the (element-wise) loss function with respect to the fitted label is calculated. As such, Newton boosting amounts to a weighted least-squares regression problem at each iteration, which is solved using base learners (e.g., using CART). As a consequence, Newton boosting can be understood as an iterative empirical risk minimization procedure in function space, that determines both the step direction and step length at the same time. This is where \texttt{XGBoostLSS} makes the connection to GAMLSS, as empirical risk minimization and Maximum Likelihood estimation are closely related. Recall from Section \ref{sec:gamlss} that GAMLSS are estimated using the first and second order partial derivatives of the log-likelihood function with respect to the distributional parameter $\theta_{k}$ of interest. By selecting an appropriate loss, or equivalently, a log-likelihood function, Maximum Likelihood can be formulated as empirical risk minimization so that the resulting XGBoost model can be interpreted as a statistical model.\footnote{Note that maximizing the negative log-likelihood is equivalent to minimizing an empirical risk function.} Besides its close relation to GAMLSS with respect to its estimation, the fact that XGBoost and \texttt{XGBoostLSS} are based on Newton boosting is also one reason for the high prediction accuracy. In a recent paper, \citet{Sigrist.2019} provides empirical evidence that Newton Boosting generally outperforms gradient boosting on the majority of data sets used for the comparison. \citet{Sigrist.2019} mainly attributes the advantage of Newton over gradient boosting to the variability in $h_{i}$, i.e., the more variation there is in the second order terms, the more pronounced is the difference between the two approaches and the more likely is Newton to outperform gradient boosting.\footnote{Also note that if $h_{i}$ is 1 everywhere, Newton and gradient boosting are equivalent. This is the case for, e.g., the squared error loss (hence assuming a Normal distribution), i.e., $l[y_{i}, \hat{y}^{(t)}_{i}] = \frac{1}{2}(\hat{y}^{(t)}_{i} - y_{i})^{2}$, we get $g_{i} = (\hat{y}^{(t)}_{i} - y_{i} )$ and $h_{i}$ = 1. As a consequence, if we use any loss function other than squared error loss, Newton tree boosting should outperform gradient boosting.}
	
	Now that we have outlined that \texttt{XGBoostLSS} can be interpreted as a statistical model by having established the connection between the estimation of GAMLSS and XGBoost, we can introduce \texttt{XGBoostLSS} more formally. Algorithm \eqref{alg:xgboostlss} gives a conceptual overview of the steps involved to estimate our model. 
	
	\begin{algorithm}[h!]
		\caption{XGBoostLSS} \label{alg:xgboostlss}
		\hspace*{\algorithmicindent} \textbf{Input:} Data set $D$ \\ 
		\hspace*{\algorithmicindent} \textbf{Required:} Appropriate (log)-likelihood/loss function $\ell[\cdot]$ \\
		\hspace*{\algorithmicindent} \textbf{Ensure:} Negative Gradient and negative Hessian exist and are non-zero
		\begin{algorithmic}[1]%
			%		\Require %
			%		\Ensure %
			\State \textbf{Step 1:} Estimate distributional parameter $\theta_{i,k}$ independently of other parameters $\theta_{i,-k}$.
			\Indent
			\For{$k$-th distributional parameter $\theta_{k}, k = 1, \ldots, K$}	%	
			\State Initialize $\hat{\theta}_{-k} = \argmax_\theta \ln \ell[\mathbf {y}, \theta_{-k}]$  \Comment{Initialize with unconditional ML-estimate} 		%	
			\State Define loss function $\ell[y, \hat{f}_{\theta_{k}}, \hat{g}_{\theta_{k}}, \hat{h}_{\theta_{k}}]$%_{\hat{\theta}_{0,-k}}$
			\State Define evaluation metric $\xi[y, \hat{f}_{\theta_{k}}]$%_{\hat{\theta}_{0,-k}}$
			\For{$m = 1, \ldots, M$ boosting iterations}	%			
			\State $\hat{g}^{m}_{\theta_{k}} = - \left[ \frac{\partial \ell[y, f(x)]}{\partial f(x)} \right]_{f(x) = \hat{f}^{(m-1)}_{\theta_{k}}(x)}$ \Comment{Negative Gradient of $k$-th parameter}%						
			\State $\hat{h}^{m}_{\theta_{k}} = - \left[ \frac{\partial^{2} \ell[y, f(x)]}{\partial f(x)^{2}} \right]_{f(x) = \hat{f}^{(m-1)}_{\theta_{k}}(x)}$ \Comment{Negative Hessian of $k$-th parameter}% 						 
			\State Determine structure of the tree% by selecting splits that optimize the split criterion.%					
			\State Estimate leaf weights				%	
			\State Update estimate: $\hat{f}^{(m)}_{\theta_{k}} = \eta \hat{f}^{(m-1)}_{\theta_{k}}$ \Comment{$\eta$ denotes the learning rate}% 
			\EndFor	
			\State Output: $\hat{f}_{\theta_{k}} = \sum^{M}_{m=0}\hat{f}^{(m)}_{\theta_{k}}, \quad k = 1, \ldots, K.$
			\EndFor 
			\EndIndent	
			\State \textbf{Step 2:} Using estimated models of Step 1, update $\hat{\theta}_{k}$ with information from $\hat{\theta}_{-k}$.
			\Indent
			\While{diff $\geq \epsilon$ and $q$ $\leq$ max\_iter} %
			\For{$k$-th distributional parameter $\theta_{k}$, $k = 1, \ldots, K$}	%
			\State Repeat steps 6-12 and update $\theta_{k}$ by incorporating information from all other parameters $\theta_{-k}$:				
			\State \begin{equation}
			\begin{aligned}
			\left(\hat{\theta}^{(q)}_{1}, \ldots, \hat{\theta}^{(q)}_{K}\right)  \quad  & \xrightarrow[]{\text{Update $\theta_{1}$}} \quad \hat{\theta}^{(q+1)}_{1} & \xrightarrow[]{\text{Output}} \quad \hat{f}^{*(q+1)}_{\theta_{1}}, \nonumber \\
			\left(\hat{\theta}^{(q+1)}_{1}, \hat{\theta}^{(q)}_{2}, \ldots, \hat{\theta}^{(q)}_{K}\right) \quad  & \xrightarrow[]{\text{Update $\theta_{2}$}} \quad \hat{\theta}^{(q+1)}_{2} & \xrightarrow[]{\text{Output}} \quad \hat{f}^{*(q+1)}_{\theta_{2}}, \nonumber \\
			\left(\hat{\theta}^{(q+1)}_{1}, \hat{\theta}^{(q+1)}_{2}, \ldots, \hat{\theta}^{(q)}_{K}\right) \quad  &  \xrightarrow[]{\text{Update $\theta_{3}$}} \quad \hat{\theta}^{(q+1)}_{3}  & \xrightarrow[]{\text{Output}} \quad \hat{f}^{*(q+1)}_{\theta_{3}}, \nonumber \\
			\quad &\makebox[\widthof{${}\xrightarrow{}$}][c]{\vdots} \\
			\left(\hat{\theta}^{(q+1)}_{1}, \hat{\theta}^{(q+1)}_{2}, \ldots, \hat{\theta}^{(q)}_{K}\right) \quad & \xrightarrow[]{\text{Update $\theta_{K}$}} \quad \hat{\theta}^{(q+1)}_{K}  & \xrightarrow[]{\text{Output}} \quad \hat{f}^{*(q+1)}_{\theta_{K}}. \nonumber
			\end{aligned}
			\end{equation}					
			\EndFor	
			%\State Output: $\hat{f}^{*(q)}_{\theta_{k}}(x)$
			\State $\mbox{deviance}_{q} \gets -2\ln \ell[\hat{f}^{*(q+1)}_{\theta} \,;\mathbf {y}]$
			\State $\mbox{diff} \gets |(\mbox{deviance}_{q+1}\ - \mbox{deviance}_{q})| / \mbox{deviance}_{q}$
			\State $q \gets q + 1$
			\EndWhile
			\EndIndent
		\end{algorithmic}
		\hspace*{\algorithmicindent} \textbf{Final Output: } $\hat{f}^{*}_{\theta_{k}}, \quad k = 1, \ldots, K.$ %= \sum^{M}_{m=0}\hat{f}^{*(m)}_{\theta_{k}}(x)
	\end{algorithm}
	
	We have designed \texttt{XGBoostLSS} in such a way that the initial XGBoost implementation remains unchanged, so that its full functionality, i.e., estimation speed and accuracy, is still available. In a sense, \texttt{XGBoostLSS} is a wrapper around XGBoost, where we interpret the loss function from a statistical perspective by formulating empirical risk minimization as Maximum Likelihood estimation. As outlined in Algorithm \eqref{alg:xgboostlss}, we first need to specify an appropriate log-likelihood, from which Gradients and Hessians are derived, that represent the partial first and second order derivatives of the log-likelihood with respect to the distributional parameter $\theta_{k}$ of interest. In contrast, however, to the approach in \citep{Mayr.2012, Thomas.2018}, that uses a component-wise gradient descent algorithm, where each of the $\theta_{k}$ is updated successively in each iteration, using the current estimates of the other distribution parameters $\theta_{-k}$ as input, our approach is a two-step procedure. In the first step, we estimate a separate model for each distributional parameter $\theta_{k}, k = 1, \ldots K$, where the unconditional Maximum Likelihood estimates of the parameters $\theta_{-k}$, not currently being estimated, are used as offset values. As such, while $\theta_{k}$ is estimated, $\theta_{-k}$ are treated being constant. Once all $\theta_{k}$ are estimated, we update each parameter by incorporating information from all other parameters until a stopping criterion based on the global deviance is met. While Step 2 is an updating of an already trained model, hyper-parameter tuning of \texttt{XGBoostLSS} is done in Step 1. For this, we use Bayesian Optimization implemented in the $\it{mlrMBO}$ package of \citet{Bischl.2017}.\footnote{In contrast to \citep{Mayr.2012, Thomas.2018}, the sequential nature of model based-optimization used in \texttt{XGBoostLSS} renders any alternating updating of the distribution parameters $\theta_{k}$ at each iteration difficult.} Once all parameters are updated and the global deviance has converged, we can draw random samples from the predicted distribution that allows us to create probabilistic forecasts from which prediction intervals and quantiles of interest can be derived. The fact that we gain insight into the data generating process, for each of the distributional parameter separately, makes \texttt{XGBoostLSS} a powerful tool.

	\section{Applications} \label{sec:applications} 
	
	In the following, we present both a simulation study and real world examples that demonstrate the functionality of \texttt{XGBoostLSS}.
	
	\subsection{Simulation}
	
	We start with a simulated a data set that exhibits heteroskedasticity, where the interest lies in predicting the 5\% and 95\% quantiles.\footnote{For the simulation, we slightly modify the example presented in \citet{Hothorn.2018a}.} The dots in red show points that lie outside the 5\% and 95\% quantiles, which are indicated by the black dashed lines.
	
	\begin{figure}[h!]
		\centering
		\includegraphics[width=0.7\linewidth]{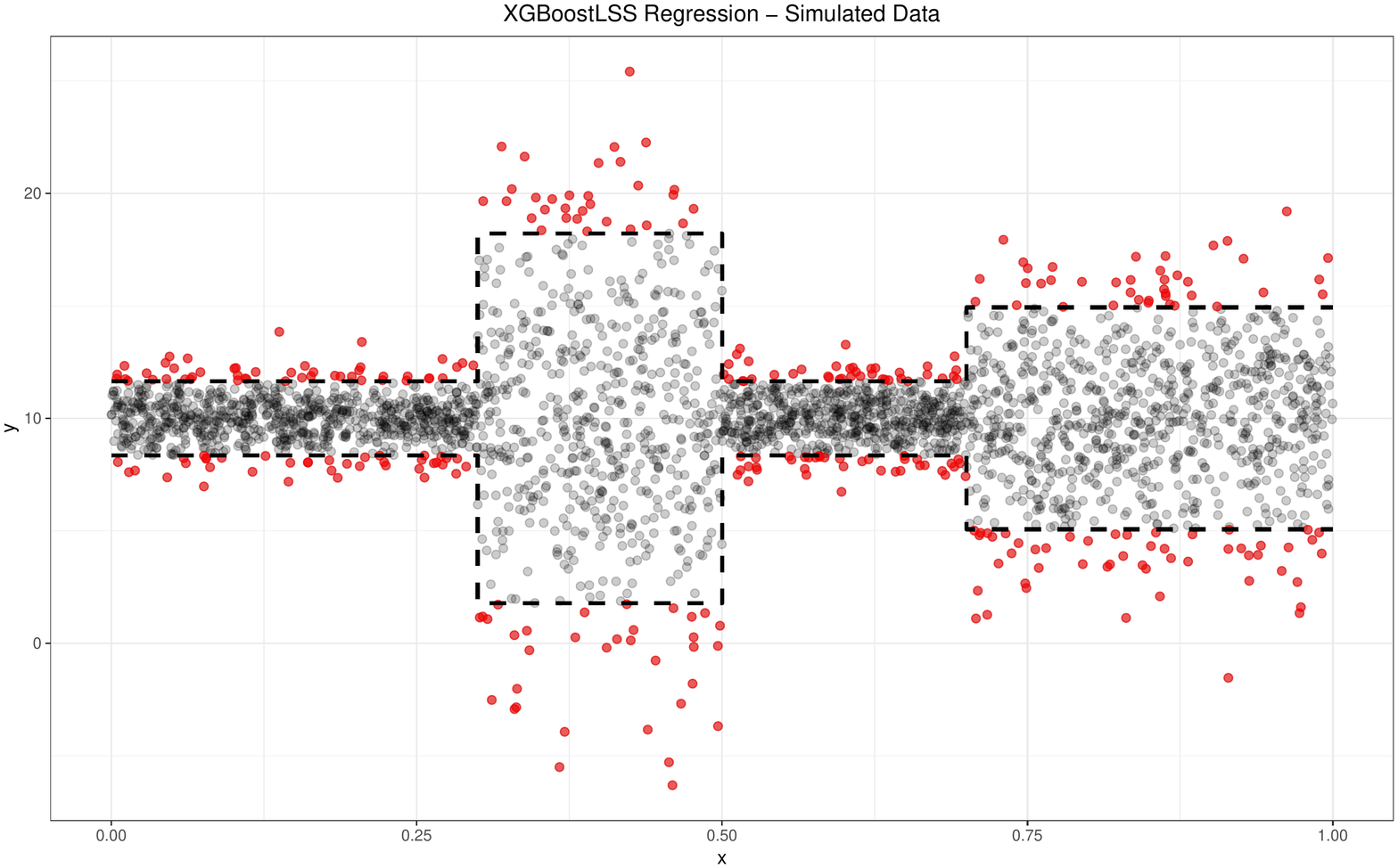}
		\caption{Simulated Train Dataset with 7,000 observations $y \sim \mathcal{N}(10,(1 + 4(0.3 < x < 0.5) + 2(x > 0.7)^{2})$. Points outside the 5\% and 95\% quantile are coloured in red. The black dashed lines depict the actual 5\% and 95\% quantiles. Besides the only informative predictor $x$, we have added $X_{1}, \ldots, X_{10}$ as noise variables to the design matrix.}
		\label{fig:sim_data}
	\end{figure}
	
	\newpage
	
	\noindent As splitting procedures, that are internally used to construct trees, can detect changes in the mean only, standard implementations of machine learning models are not able to recognize any distributional changes (e.g., change of variance), even if these can be related to covariates \citep{Hothorn.2018a}. As such, XGBoost doesn't provide any uncertainty quantification in its current implementation, as the model focuses on predicting the conditional mean $\mathbb{E}(Y|\mathbf{X} = \mathbf{x})$ only, without any assessment on the full predictive distribution $F_{Y}(y|\mathbf{x})$. This is in contrast to \texttt{XGBoostLSS}, where all distributional parameters are modelled as functions of covariates.  
	
\begin{figure}[h!]
		\centering
		\includegraphics[width=0.8\linewidth]{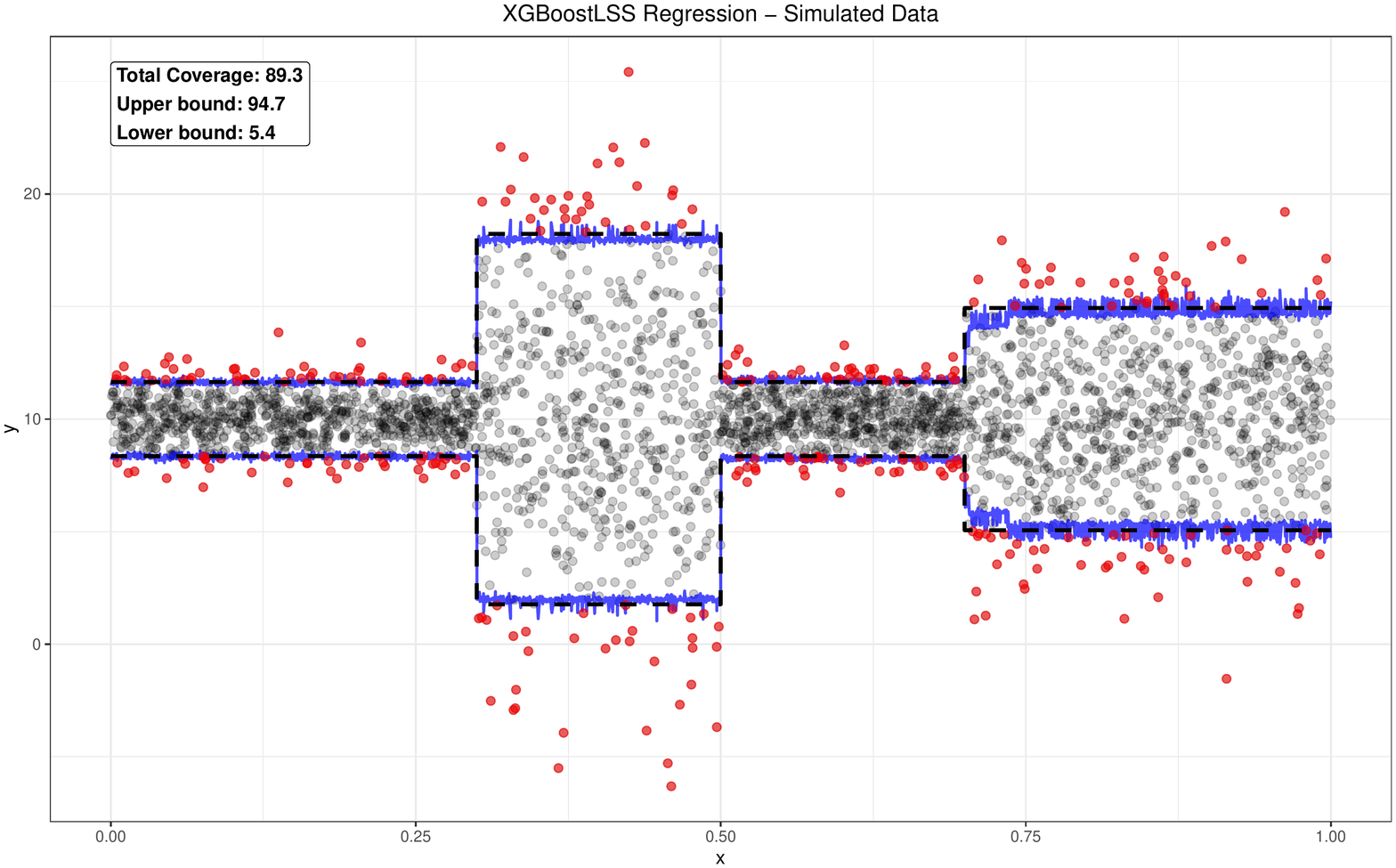}
		\caption{Simulated Test Dataset with 3,000 observations $y \sim \mathcal{N}(10,(1 + 4(0.3 < x < 0.5) + 2(x > 0.7)^{2})$. Points outside the conditional 5\% and 95\% quantile are in red. The black dashed lines depict the actual 5\% and 95\% quantiles. Conditional 5\% and 95\% quantile predictions obtained from \texttt{XGBoostLSS} are depicted by the blue lines. Besides the only informative predictor $x$, we have added $X_{1}, \ldots, X_{10}$ as noise variables to the design matrix.}
		\label{fig:sim_mbo}
	\end{figure}

	Let's fit our \texttt{XGBoostLSS} model to the data. In general, the syntax is similar to the original XGBoost implementation. However, the user has to make a distributional assumption by specifying a family in the function call. As the data has been generated by a Normal distribution, we use the Normal as a function input. The user also has the option of providing a list of hyper-parameters that are used for training the surrogate regression model to find an optimized set of parameters.\footnote{Currently, the default set-up in \texttt{XGBoostLSS} optimizes $eta, gamma, max\_depth, min\_child\_weight, subsample$ and $colsample\_bytree$ as hyper-parameters.} Once the model is trained, we can predict all parameters of the distribution. As \texttt{XGBoostLSS} allows to model the entire conditional distribution, we obtain prediction intervals and quantiles of interest directly from the predicted quantile function. Figure \ref{fig:sim_mbo} shows the predictions of \texttt{XGBoostLSS} for the 5\% and 95\% quantile in blue.
	% we can draw random samples from the predicted distribution, which allows us to create prediction intervals and quantiles of interest.
	
	\noindent Comparing the coverage of the intervals with the nominal level of 90\% shows that \texttt{XGBoostLSS} does not only correctly model the heteroskedasticity in the data, but it also provides an accurate forecast for the 5\% and 95\% quantiles. To assess its ability across the entire response distribution, Table \ref{tab:coverage} compares the coverage of \texttt{XGBoostLSS} across several quantiles.
	
	\begin{table}[h!]
		\begin{center}
			\begin{threeparttable}
				\caption{Empirical Coverage}
				\begin{tabular}{rrrrrrrr}
					\toprule
					& (5, 95) & (10, 90)  & (20, 80) & (30, 70) & (40, 60) & (50, 50) \\  
					\midrule
					Total Coverage        & 89.3    & 79.3      & 60.1     & 39.5     & 19.7     & 0 \\
					Upper Bound  		  & 94.7    & 89.6      & 79.2     & 68.7     & 59.4     & 49.9 \\ 
					Lower Bound 		  & 5.4     & 10.3      & 19.1     & 29.2     & 39.7     & 49.9 \\  
					\bottomrule
				\end{tabular}
				%			\begin{tablenotes}
				%				\tiny
				%				\item \noindent Average Continuous Ranked Probability Scoring Rules (CRPS); Average Logarithmic Score (LOG); Mean Absolute Percentage Error (MAPE); Mean Square Error (MSE); Root Mean Square Error (RMSE); Mean Absolute Error (MAE); Median Absolute Error (MEDIAN-AE); Relative Absolute Error (RAE); Root Mean Square Percentage Error (RMSPE); Root Mean Squared Logarithmic Error (RMSLE); Root Relative Squared Error (RRSE); R-Squared/Coefficient of Determination (R$^{2}$). Best out-of-sample results are marked in bold (lower is better, except $R^{2}$).
				%			\end{tablenotes}
				\label{tab:coverage}
			\end{threeparttable}
		\end{center}
	\end{table}
	
	The great flexibility of \texttt{XGBoostLSS} also comes from its ability to provide attribute importance, as well as partial dependence plots, for all of the distributional parameters. In the following we only investigate the effect on the conditional variance. All inference plots are generated using wrappers around the $\it{interpretable}$ $\it{machine}$ $\it{learning}$ $\it{(iml)}$ package of \citet{Molnar.2018}.
	
	%For ease of visualization, we also plot the mean of the predicted values for each region of constant variance in Figure \ref{fig:sim_mbo_mean}.
	%
	%\begin{figure}[h!]
	%	\centering
	%	\includegraphics[width=0.7\linewidth]{XGBoostLSS_mbo_sim_mean.pdf}
	%	\caption{Simulated Test Dataset with 3,000 observations $Y \sim N(10,(1 + 4(0.3 < x < 0.5) + 2(x > 0.7)^{2})$. Points outside the conditional 5\% and 95\% quantile are in red. The black dashed lines depict the actual 5\% and 95\% quantiles. Mean of the predicted conditional 5\% and 95\% quantile obtained from \texttt{XGBoostLSS} are depicted by the blue lines.}
	%	\label{fig:sim_mbo_mean}
	%\end{figure}
	
	\begin{figure}[h!]
		\centering
		\includegraphics[width=0.7\linewidth]{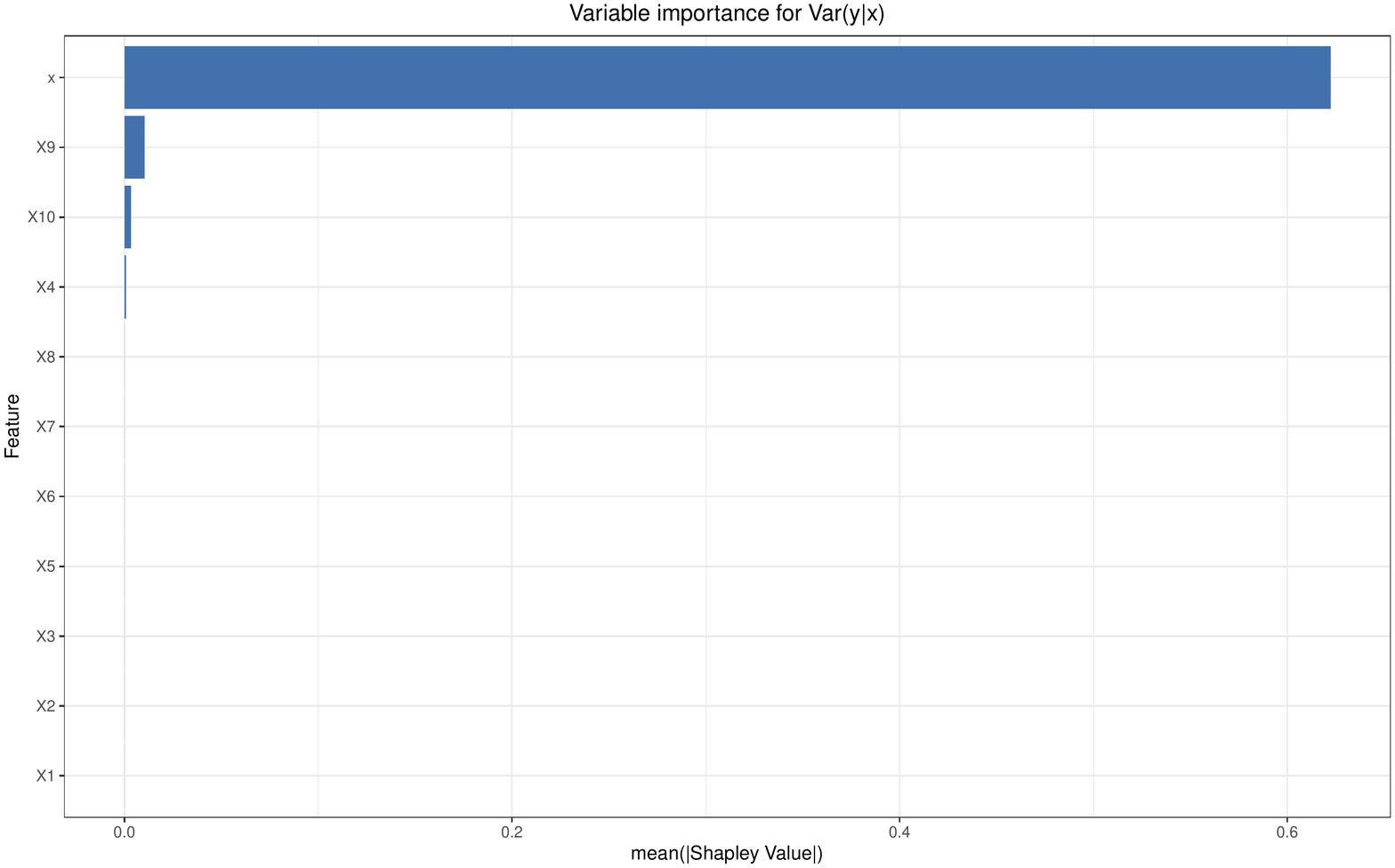}
		\caption{Mean Absolute Shapley Value of $\mathbb{V}(Y|\mathbf{X} = \mathbf{x})$.}
		\label{fig:shap_value}
	\end{figure}
	
	\noindent The plot of the Shapley value shows that \texttt{XGBoostLSS} has identified the only informative predictor $x$ and does not consider any of the noise variables $X_{1}, \ldots, X_{10}$ as important features. Looking at partial dependence plots of $\mathbb{V}(Y|\mathbf{X} = \mathbf{x})$ shows that it also correctly identifies the heteroskedasticity in the data.
	
	\begin{figure}[h!]
		\centering
		\includegraphics[width=0.7\linewidth]{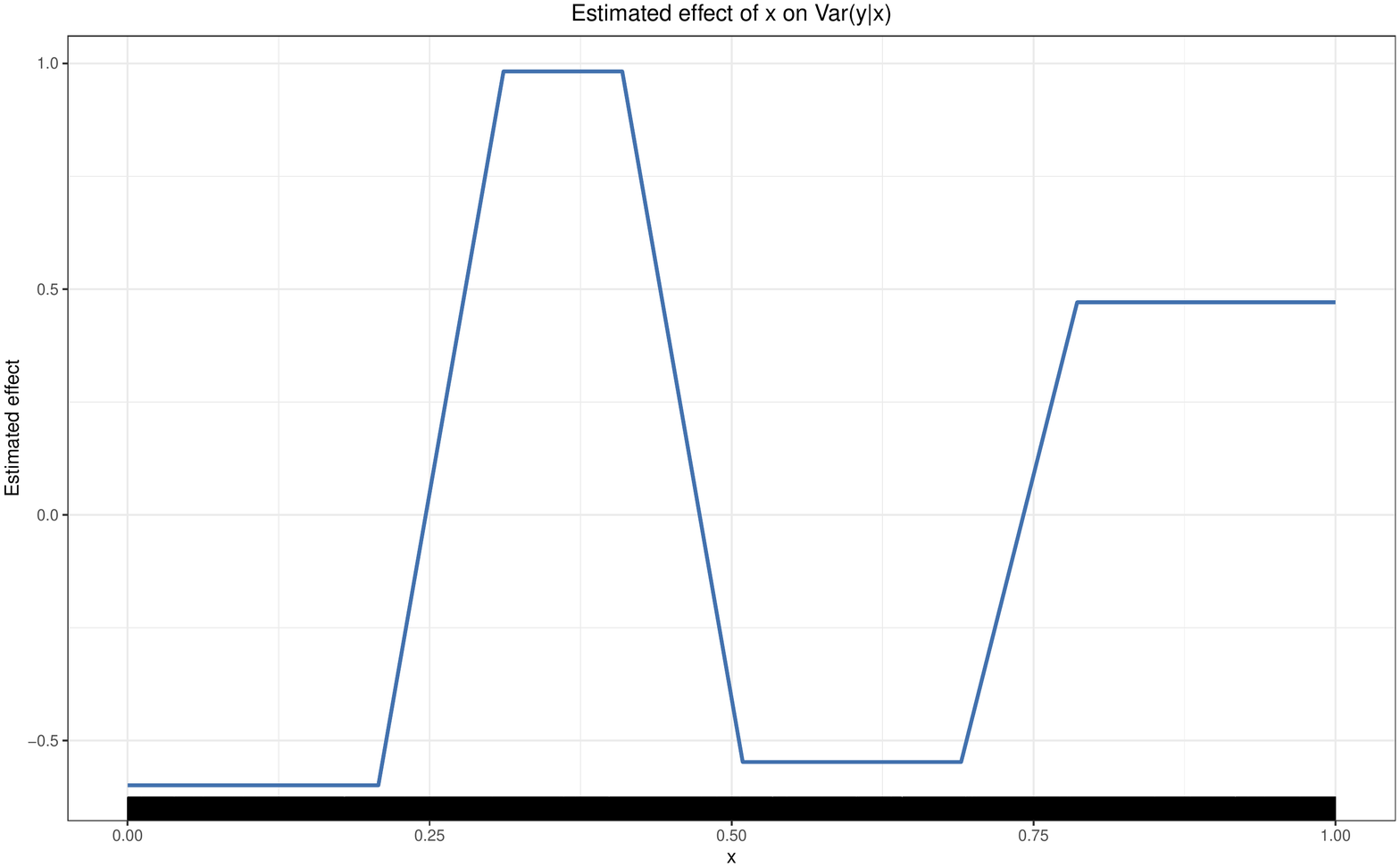}
		\caption{Partial Dependence Plot of $\mathbb{V}(Y|\mathbf{X} = \mathbf{x})$.}
		\label{fig:part_effec}
	\end{figure}
	
	\newpage
	
	\subsection{Munich Rent}
	
\begin{figure}[h!]
		\centering
		\includegraphics[width=0.6\linewidth]{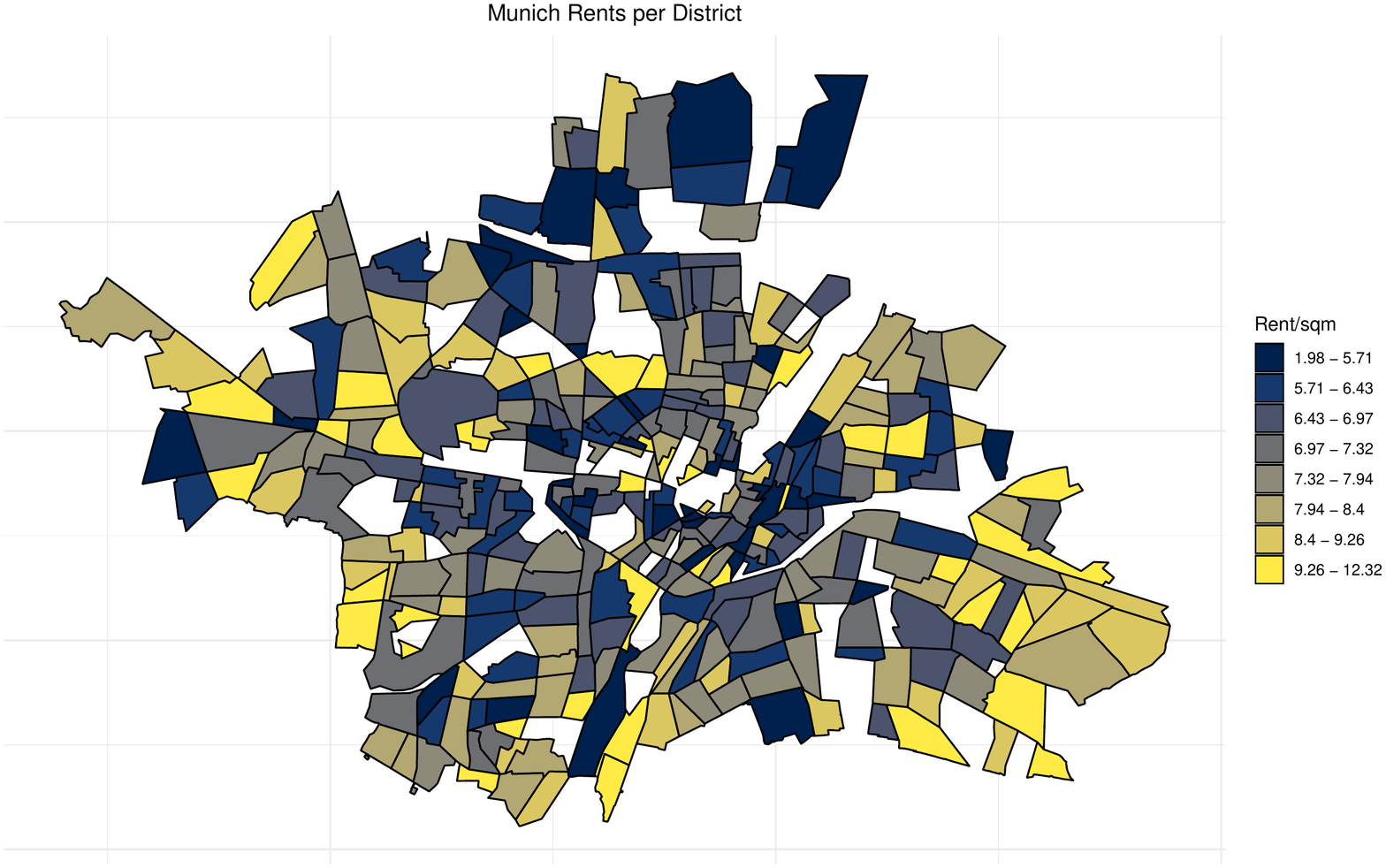}
		\caption{Munich Rents per square meter per district.}
		\label{fig:rent_map}
	\end{figure}

	Considering there is an active discussion around imposing a freeze in German cities on rents, we have chosen to re-visit the famous Munich Rent data set, as Munich as among the most expensive cities in Germany when it comes to living costs. In this example, we illustrate the functionality of \texttt{XGBoostLSS} using a sample of 2,053 apartments from the data collected for the preparation of the Munich rent index 2003, as shown in Figure \ref{fig:rent_map}. As our dependent variable, we select \emph{Net rent per square meter in EUR}.
	
	The first decision one has to make is about choosing an appropriate distribution for the response. As there are many potential candidates, we use an automated approach based on the generalised Akaike information criterion (GAIC). Due to its infrastructure and available distributions, \texttt{XGBoostLSS} relies on the package $\it{gamlss}$ of \citet{Stasinopoulos.2017}.
	
	\begin{table}[h!]
		\begin{center}
			\caption{Candidate Response Distributions}
			\scalebox{0.95}{%
				\begin{threeparttable}
					
					\begin{tabular*}{0.5\textwidth}{l @{\extracolsep{\fill}}r}
						\toprule
						Distribution  & GAIC    \\
						\midrule
						GB2     & 6588.29 \\
						NO      & 6601.17 \\
						GG      & 6602.02 \\
						BCCG    & 6602.26 \\
						WEI     & 6602.37 \\
						exGAUS  & 6603.17 \\
						BCT     & 6603.35 \\
						BCPEo   & 6604.26 \\
						GA      & 6707.85 \\
						GIG     & 6709.85 \\
						LOGNO   & 6839.56 \\
						IG      & 6871.12 \\
						IGAMMA  & 7046.50 \\
						EXP     & 9018.04 \\
						PARETO2 & 9020.04 \\
						GP      & 9020.05 \\
						\bottomrule
					\end{tabular*}
					\begin{tablenotes}
						\tiny
						\item \noindent Generalized Beta Type 2 (GB2); Normal (NO); Generalized Gamma (GG); Box-Cox Cole and Green (BCCG); Weibull (WEI); ex-Gaussian (exGAUS); Box-Cox t-distribution (BCT); Box-Cox Power Exponential (BCPEo); Gamma (GA); Generalized Inverse Gaussian (GIG); Log-Normal (LOGNO); Inverse Gaussian (IG); Inverse Gamma (IGAMMA); Exponential (EXP); Pareto Type 2 (PARETO2); Generalized Pareto (GP).
					\end{tablenotes}
			\end{threeparttable}}
			\label{tab:dist}
		\end{center}
	\end{table}
	
	\newpage
	
	\noindent Even though Table \ref{tab:dist} suggests the Generalized Beta Type 2 to provide the best approximation to the data, we use the more parsimonious Normal distribution, as it has only two distributional parameters, compared to 4 of the Generalized Beta Type 2. In general, though, \texttt{XGBoostLSS} is flexible to allow the user to choose and fit all distributions available in the $\it{gamlss}$ package. The good fit of the Normal distribution is also confirmed by the the density plot, where the response of the train data is presented as a histogram, while the fitted Normal is shown in red. 
	
	\begin{figure}[h!]
		\centering
		\includegraphics[width=0.6\linewidth]{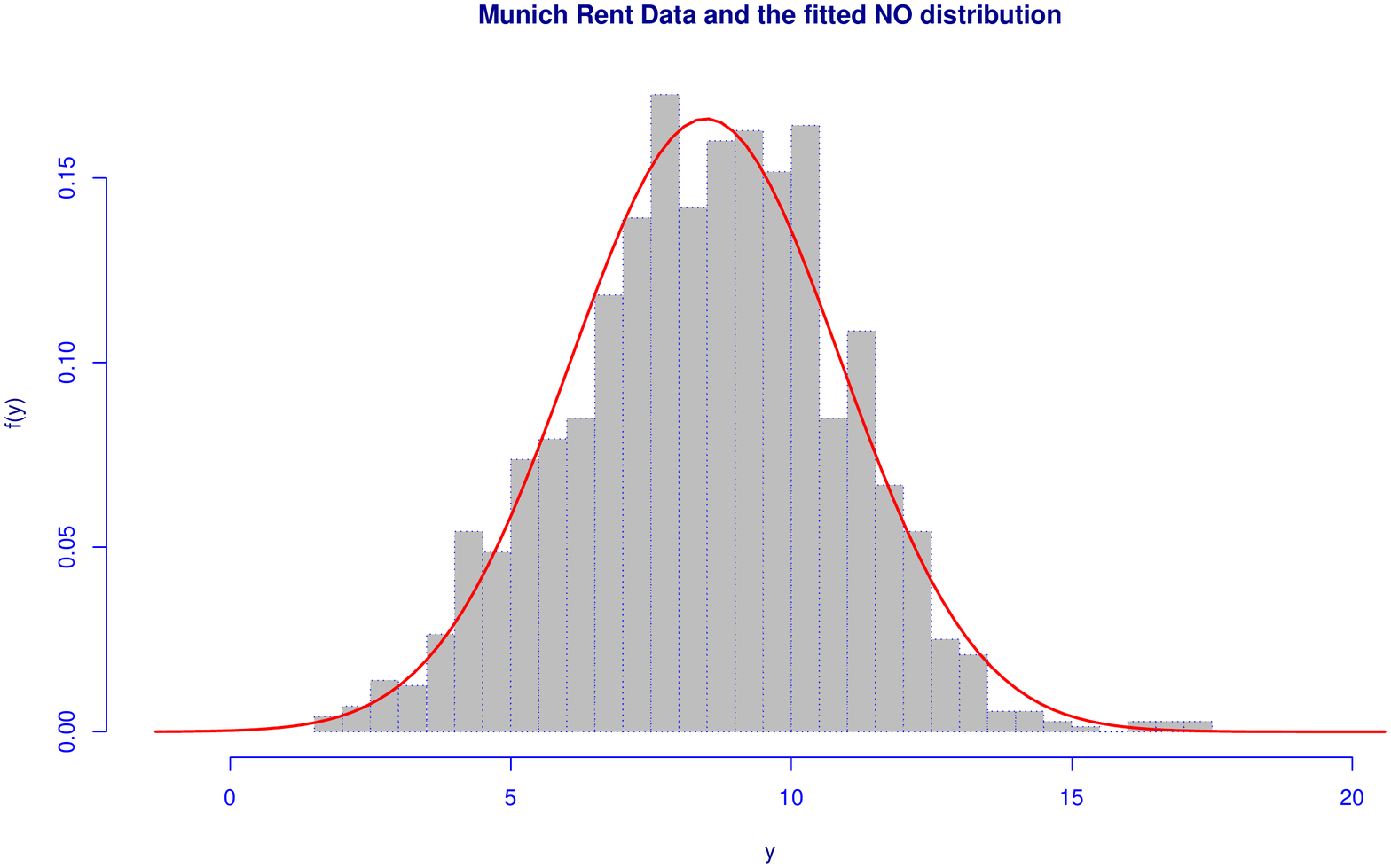}
		\caption{Fitted Normal Distribution.}
		\label{fig:fitdist}
	\end{figure}
	
\begin{figure}[h!]
		\centering
		\includegraphics[width=0.72\linewidth]{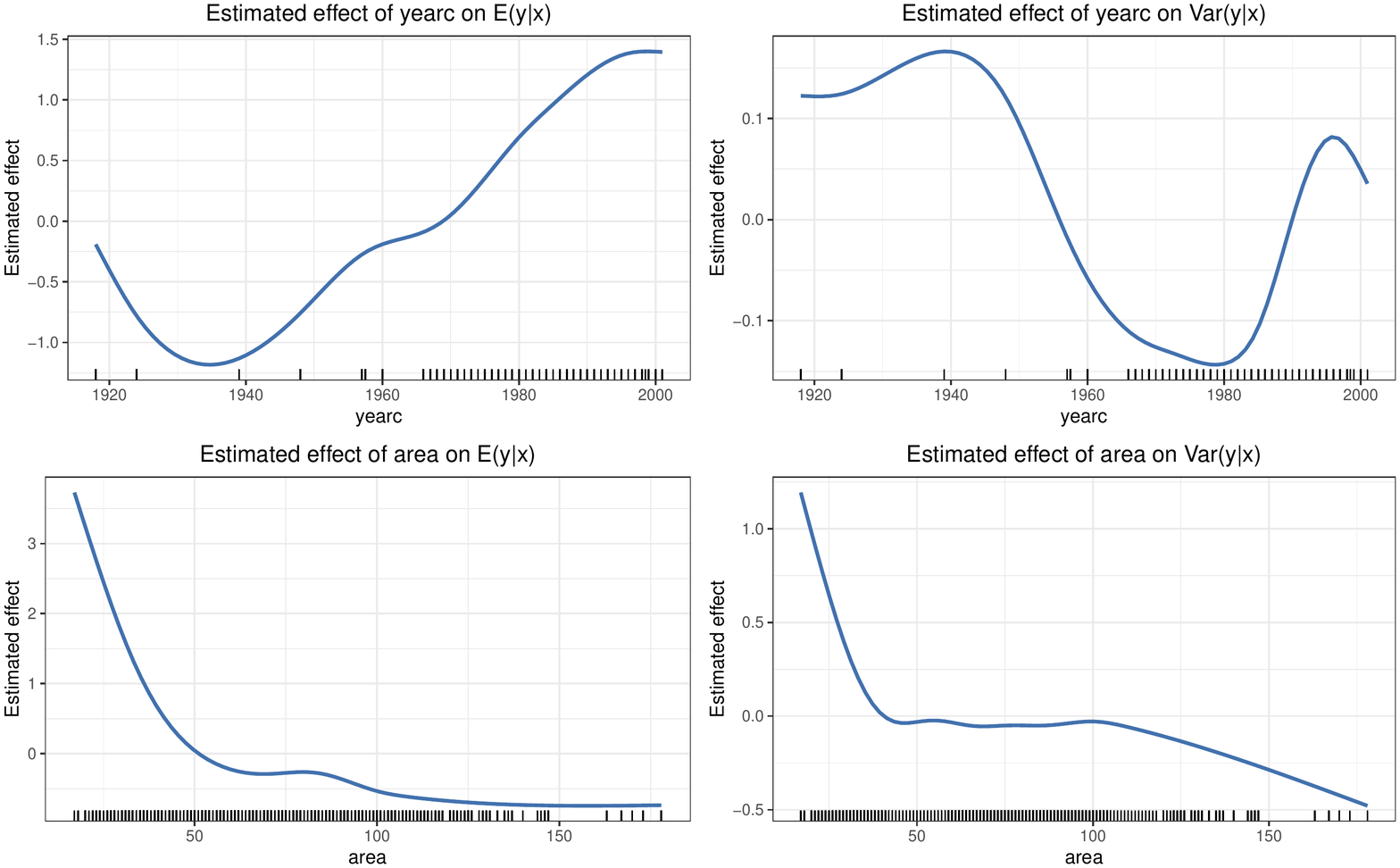}
		\caption{Estimated Partial Effects.}
		\label{fig:pdp}
	\end{figure}

	\noindent Now that we have specified the distribution, we fit our \texttt{XGBoostLSS} model to the data. Again, we use Bayesian Optimization for finding an optimal set of hyper-parameters.\footnote{In its current implementation, \texttt{XGBoostLSS} uses a time-budget parameter, that indicates the running time budget in minutes, as a stopping criteria for the Bayesian Optimization. For the Munich Rent example presented in this section, we set the time budget to 5 minutes.} Looking at the estimated effects presented in Figure \ref{fig:pdp} indicates that newer flats are on average more expensive, with the variance first decreasing and increasing again for flats built around 1980 and later. Also, as expected, rents per square meter decrease with an increasing size of the apartment.
	
	\newpage
	
	\noindent The diagnostics for \texttt{XGBoostLSS} are based on quantile residuals of the fitted model.\footnote{For continuous response data, the quantile residuals are based on $u_{i} = F_{i}(y_{i}|\hat{\veclatin{\theta}})$, where $F_{i}(\cdot)$ is the cumulative distribution function estimated for the $i$-th observation, $\hat{\veclatin{\theta}}$ contains all estimated parameters and $y_{i}$ is the corresponding observation. If $F_{i}(\cdot)$ is close to the true distribution of $y_{i}$, then $u_{i}$ approximately follows a uniform distribution. The quantile residuals are then defined as $\hat{r}_{i} = \phi^{-1}(u_{i})$, where $\phi^{-1}(\cdot)$ is the inverse cumulative distribution function of the standard Normal distribution. Hence, ${r}_{i}$ is approximately standard Normal if the estimated model is close to the true one.}	
	
	\begin{figure}[h!]
		\centering
		\includegraphics[width=0.65\linewidth]{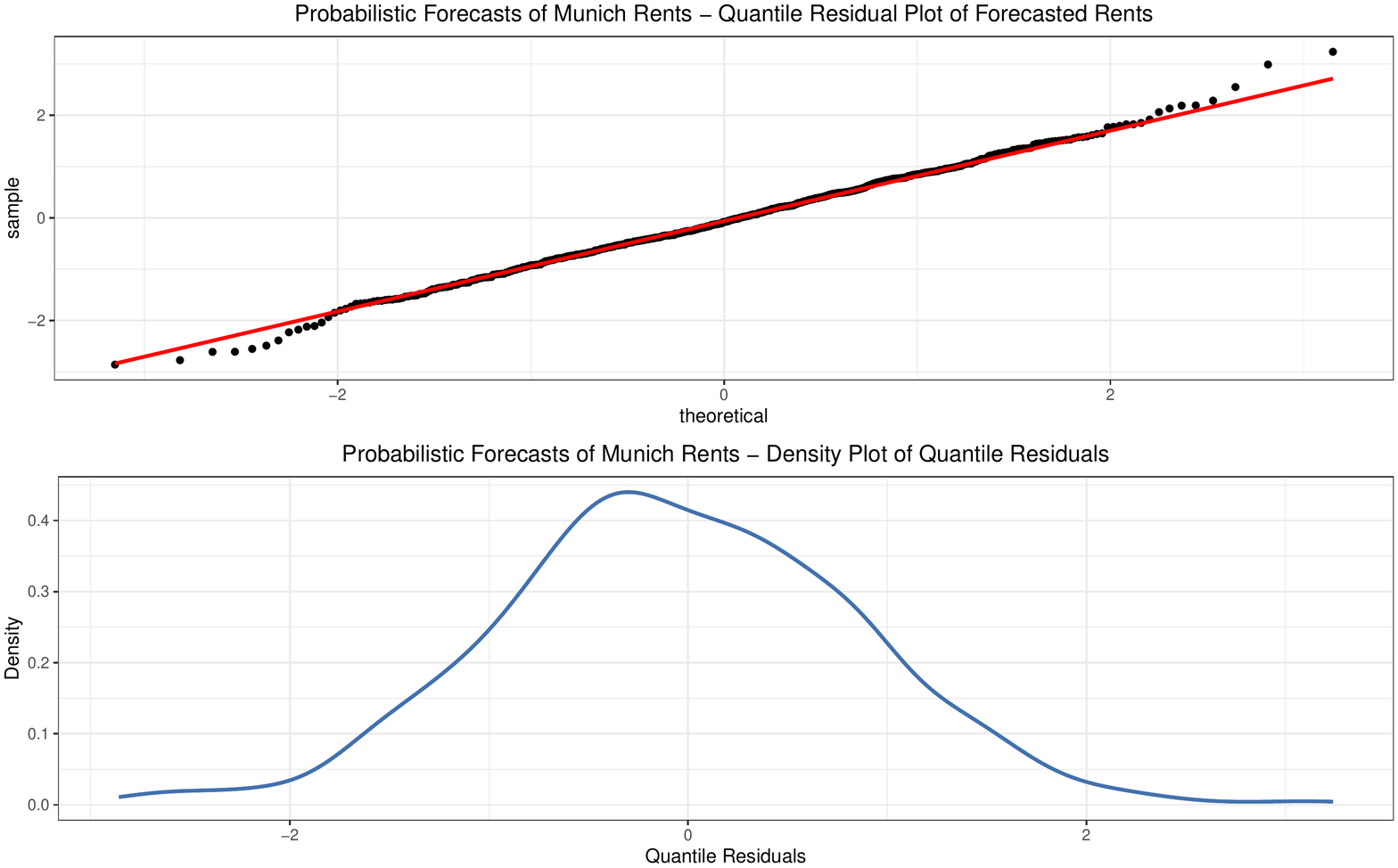}
		\caption{Quantile Residuals.}
		\label{fig:quantres}
	\end{figure}
	
	\noindent Despite some slight under-fitting in the tails of the distribution, \texttt{XGBoostLSS} provides a well calibrated forecast and confirms that our model is a good approximation to the data. \texttt{XGBoostLSS} also allows to investigate feature importances for all distributional parameters. Looking at the top 10 Shapley values for both the conditional mean and variance indicates that both $yearc$ and $area$ are considered as being the most important variables.

	\begin{figure}[h!]
		\centering
		\includegraphics[width=0.7\linewidth]{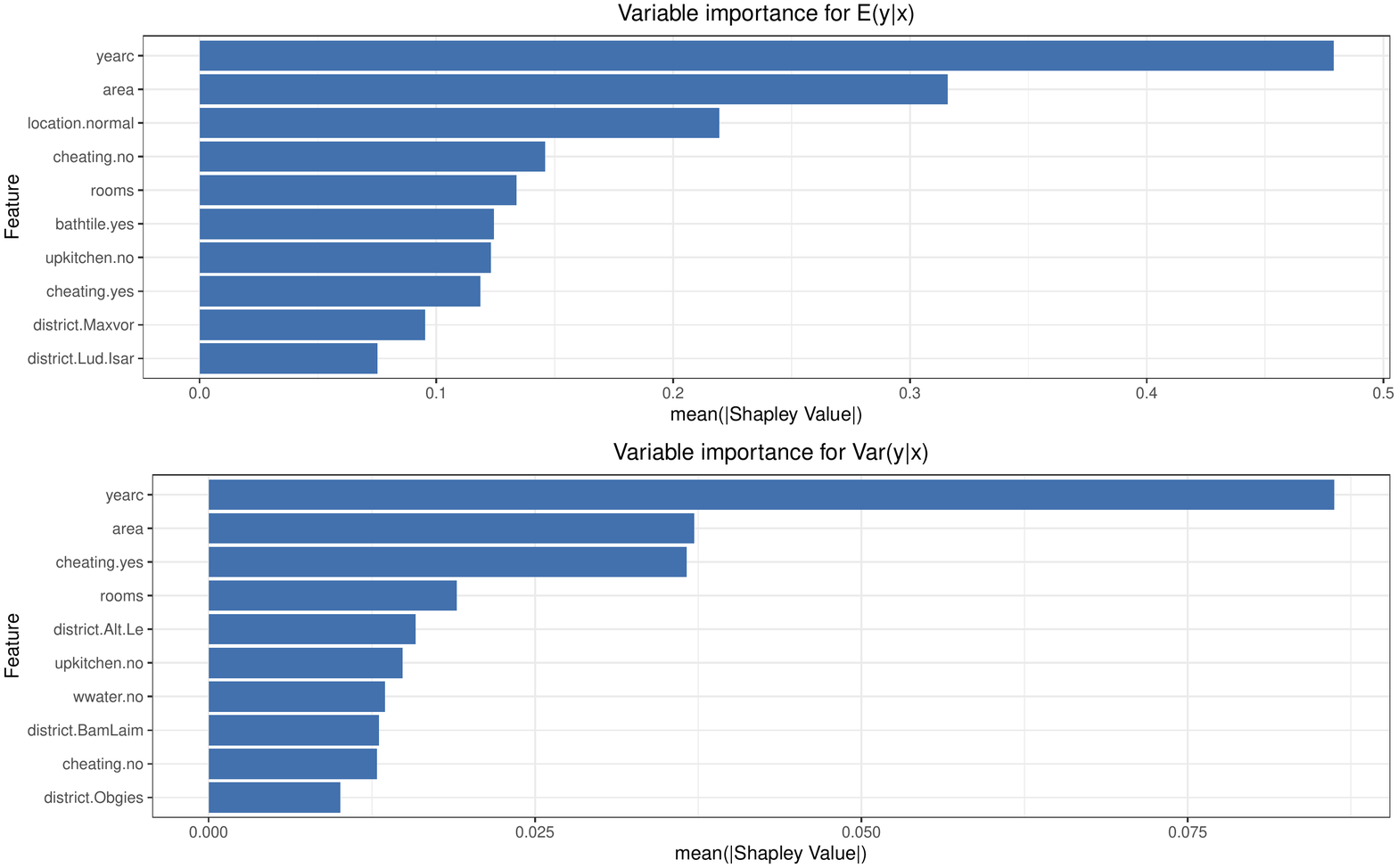}
		\caption{Mean Absolute Shapley Value of  $\mathbb{E}(Y|\mathbf{X} = \mathbf{x})$ and $\mathbb{V}(Y|\mathbf{X} = \mathbf{x})$.}
		\label{fig:munich_shap}
	\end{figure}
	
	\newpage
	
	\noindent Besides the global attribute importance, the user might also be interested in local attribute importance for each single prediction individually. This allows to answer questions like 'How did the feature values of a single data point affect its prediction?' For illustration purposes, we select the first predicted rent of the test data set and present the local attribute importance for $\mathbb{E}(Y|\mathbf{X} = \mathbf{x})$ .
	
	\begin{figure}[h!]
		\centering
		\includegraphics[width=0.7\linewidth]{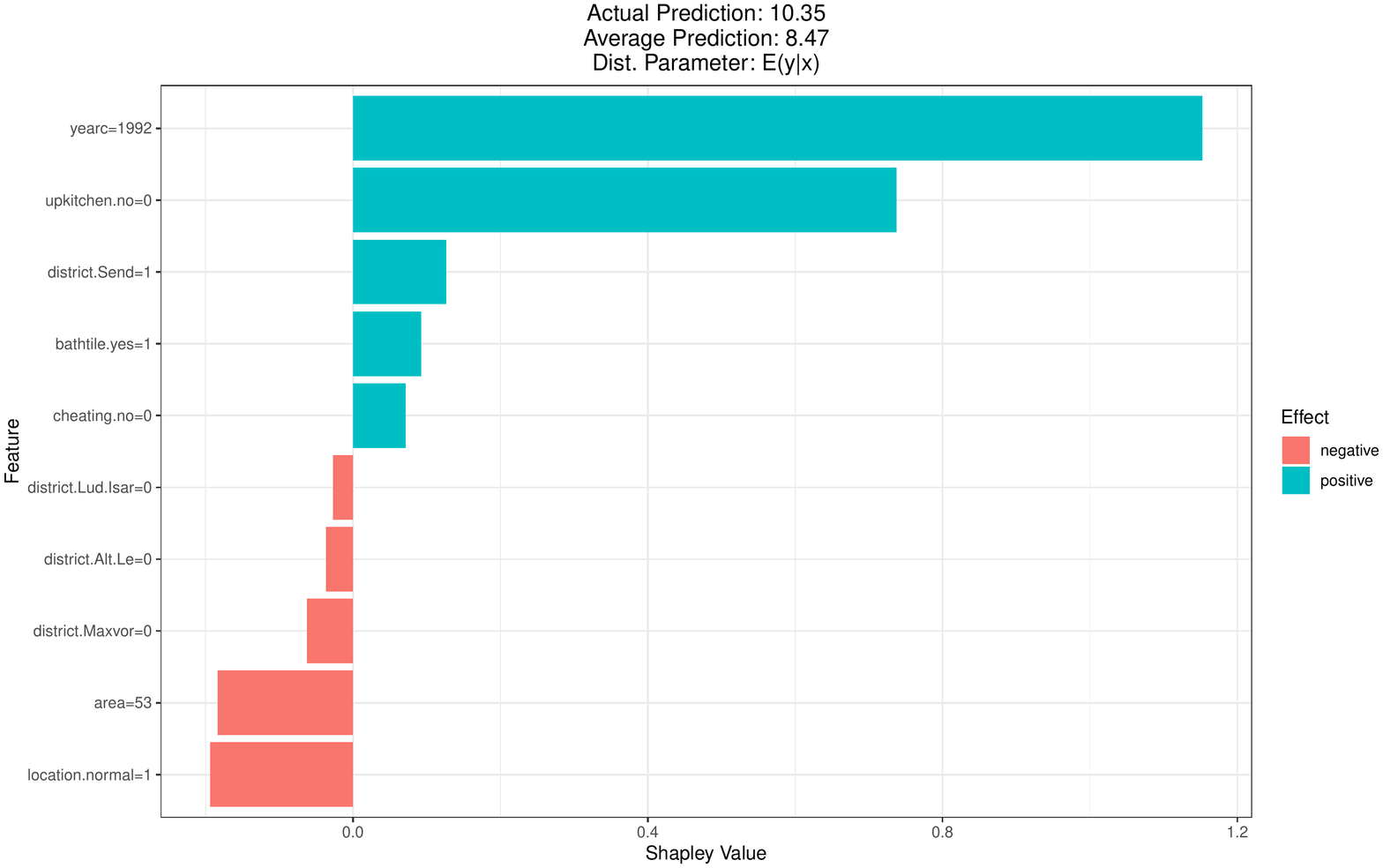}
		\caption{Local Shapley Value of $\mathbb{E}(Y|\mathbf{X} = \mathbf{x})$.}
		\label{fig:munich_shap_local}
	\end{figure}
	
	\noindent We can also measure how strongly features interact with one other. The range of the measure is between 0 (no interaction) and 1 (strong interaction).
	
	\begin{figure}[h!]
		\centering
		\includegraphics[width=0.6\linewidth]{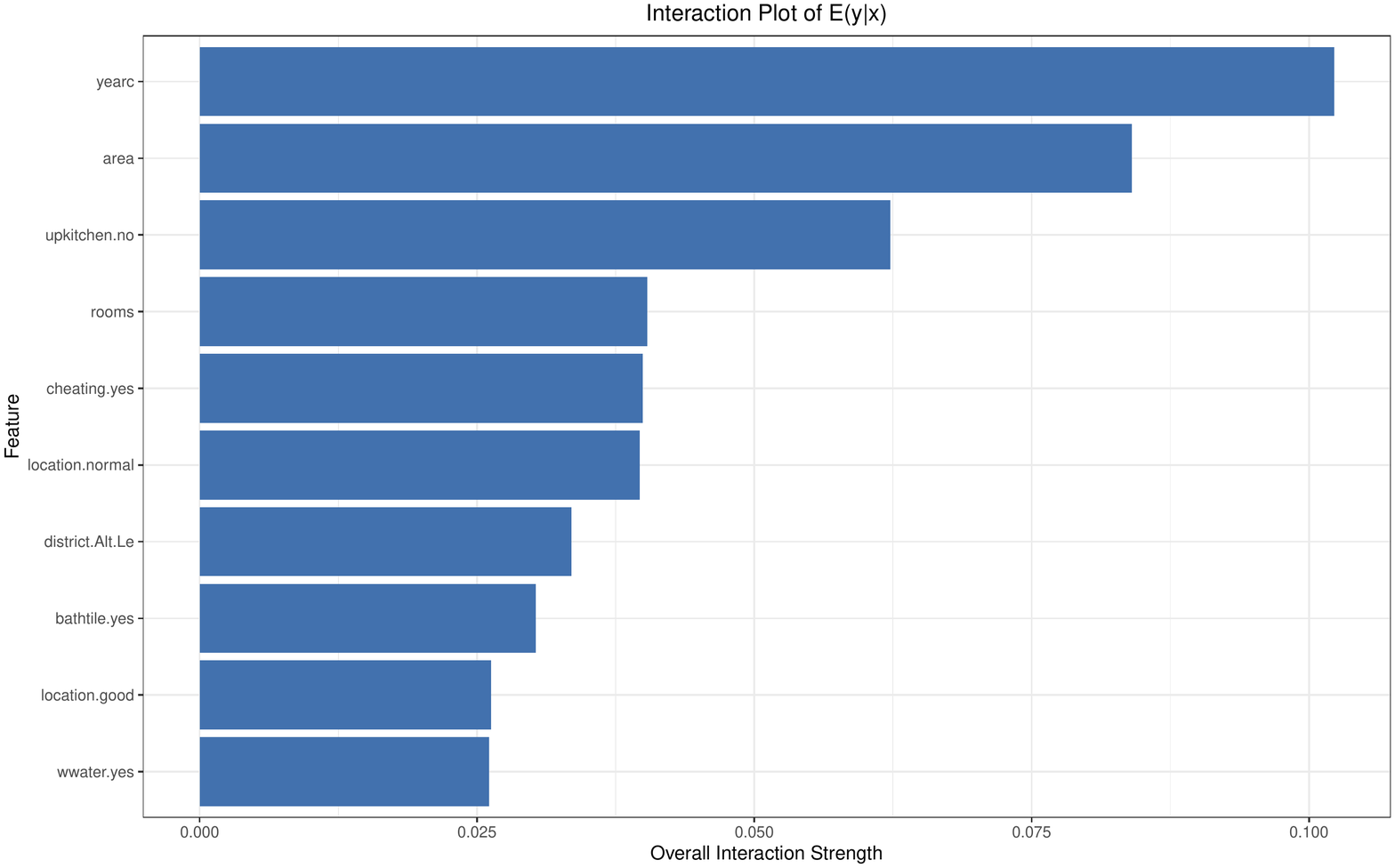}
		\caption{Interaction Plot of $\mathbb{E}(Y|\mathbf{X} = \mathbf{x})$.}
		\label{fig:munich_inter}
	\end{figure}
	
	\noindent Among all covariates, $yearc$ seems to have the strongest interaction. We can also further analyse its effect and specify a feature and measure all its 2-way interactions with all other features.
	
	\newpage
	
	\begin{figure}[h!]
		\centering
		\includegraphics[width=0.6\linewidth]{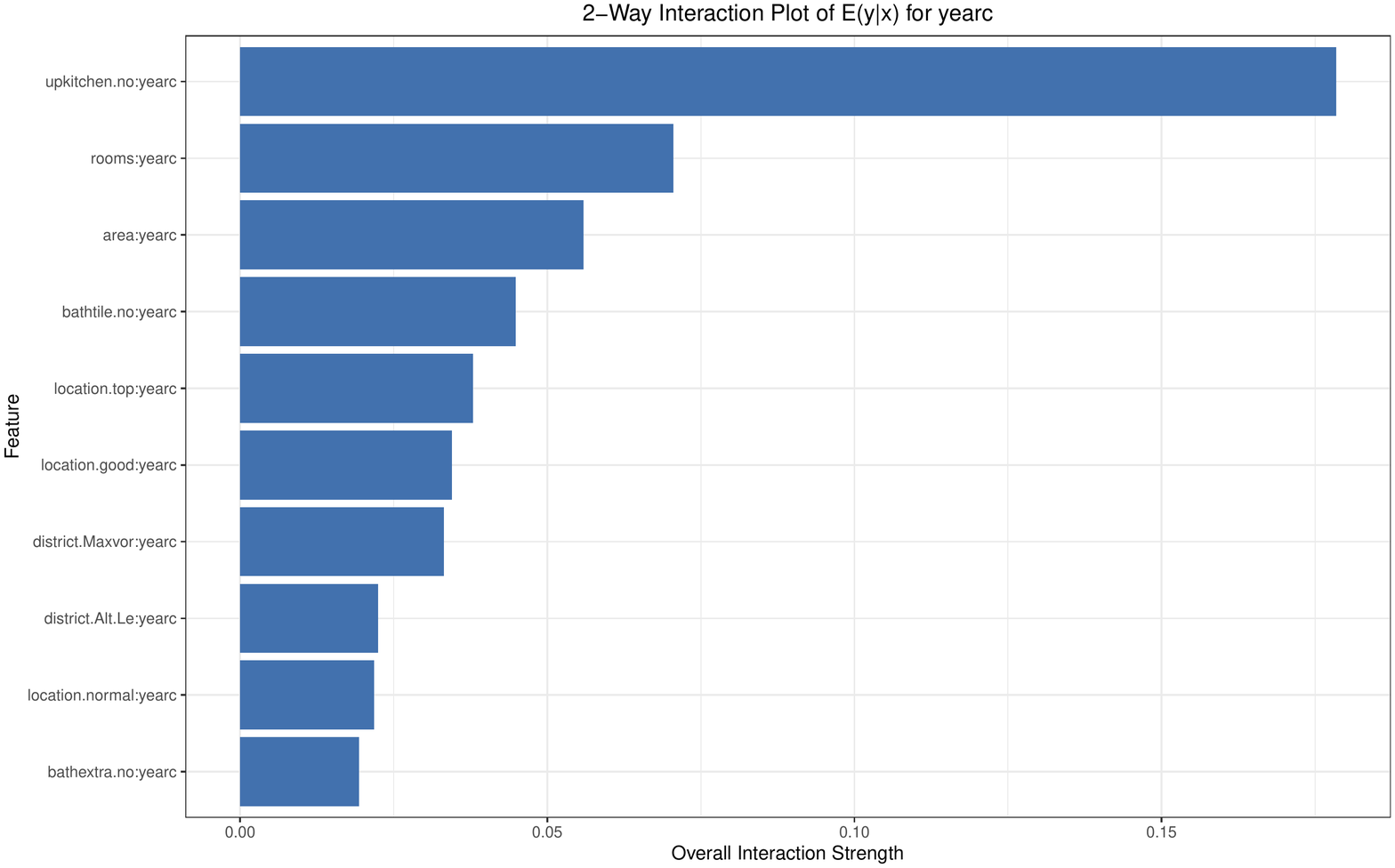}
		\caption{2-way interaction Plot for $yearc$ of $\mathbb{E}(Y|\mathbf{X} = \mathbf{x})$.}
		\label{fig:munich_inter1}
	\end{figure}
	
\begin{figure}[h!]
		\centering
		\includegraphics[width=0.8\linewidth]{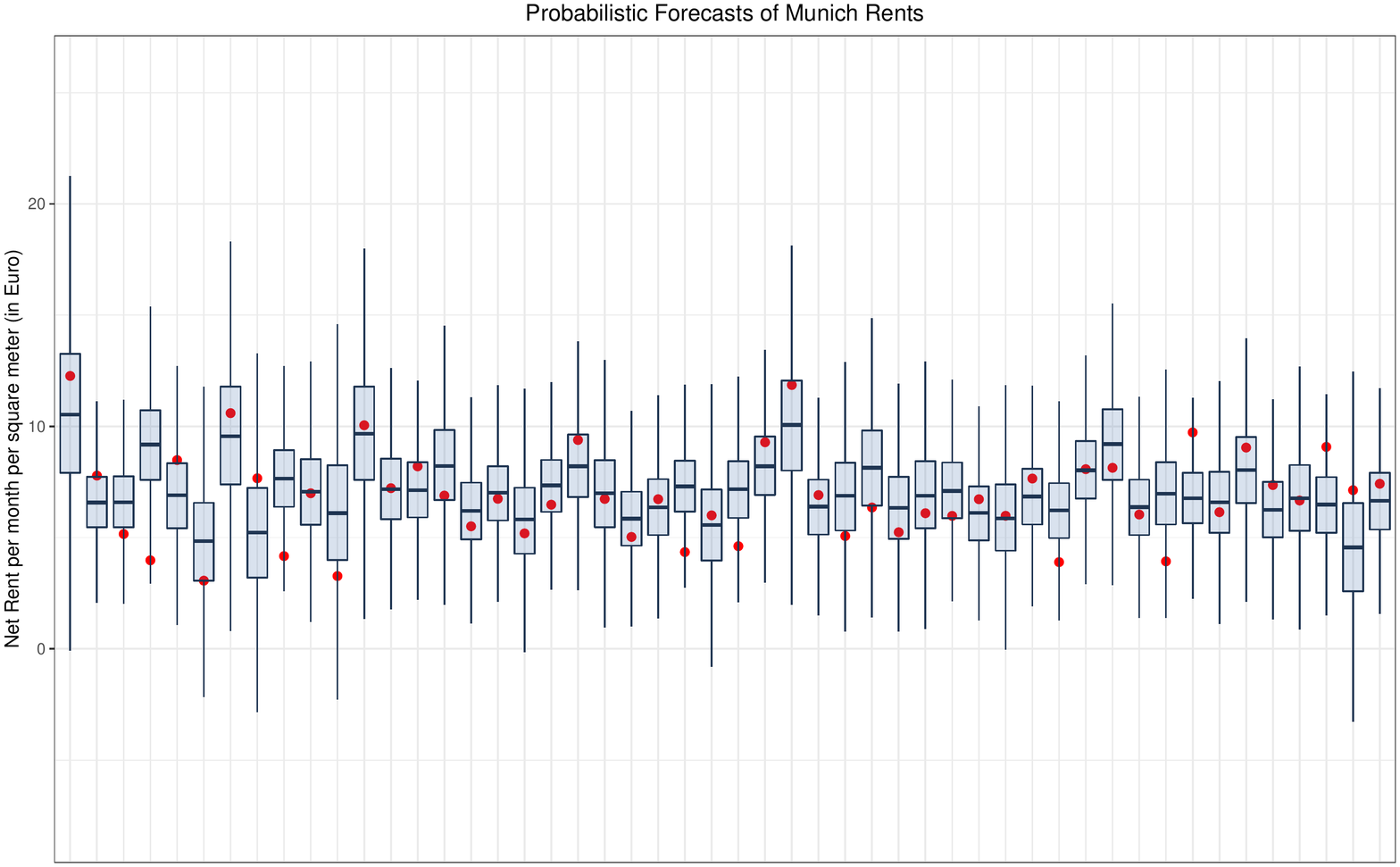}
		\caption{Boxplots of Probabilistic Forecasts of Munich Rents.}
		\label{fig:munich_forecast}
	\end{figure}

	\noindent As we have modelled all parameters of the Normal distribution, \texttt{XGBoostLSS} provides a probabilistic forecast, from which any quantity of interest can be derived. Figure \ref{fig:munich_forecast} shows a random subset of 50 predictions only for ease of readability. The red dots show the actual out of sample rents, while the boxplots visualise the predictions.
	
	\noindent Even though the Normal distribution was identified by the GAIC as an appropriate distribution, the Whiskers in Figure \ref{fig:munich_forecast} show that some of the forecasted rents are actually negative. In real life applications, a distribution with strictly positive support might be a more reasonable choice. Also, we can plot a subset of the forecasted densities and cumulative distributions.
	
	\begin{figure}[h!]
		\centering
		\includegraphics[width=0.7\linewidth]{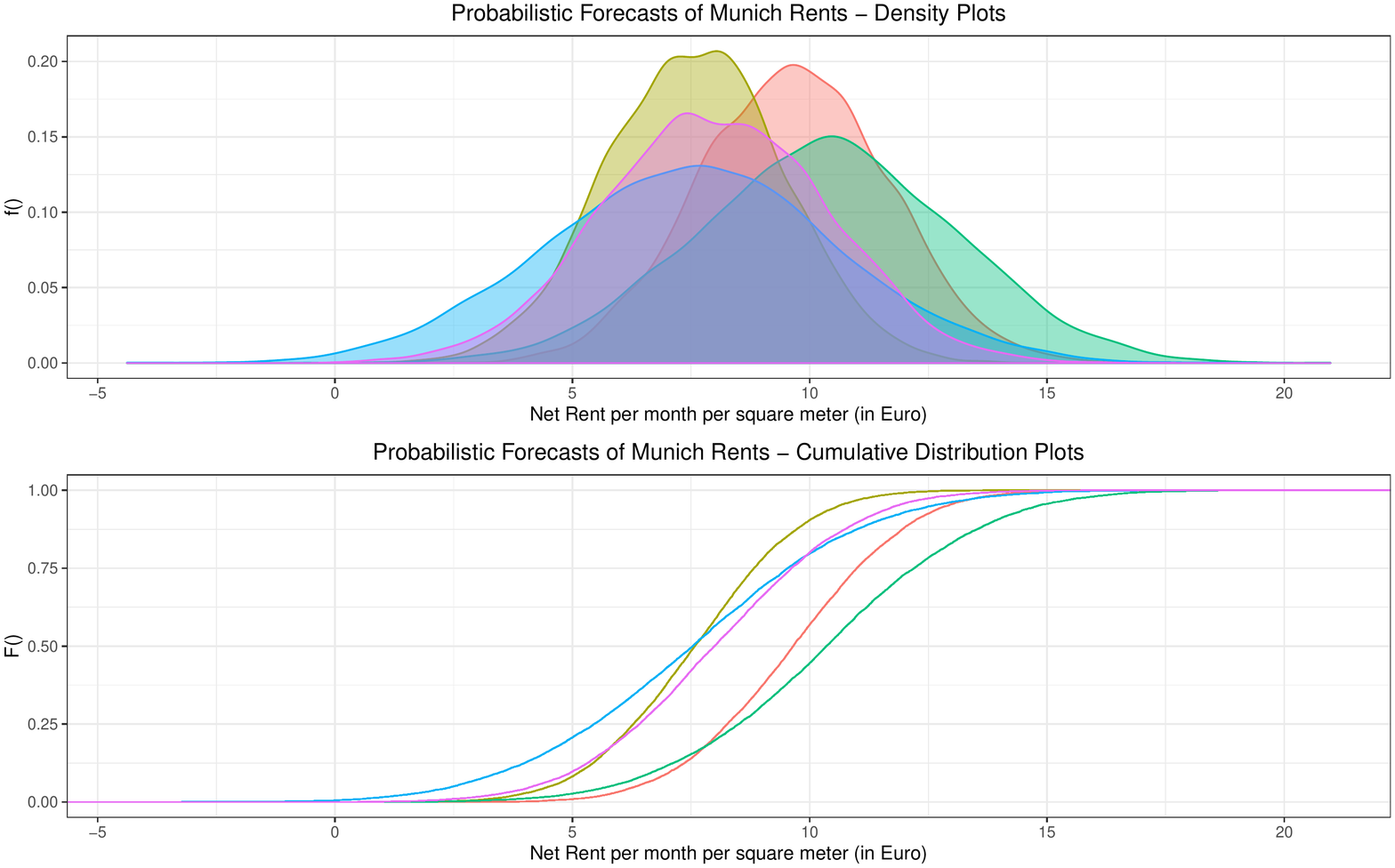}
		\caption{Density and Cumulative Distribution Plots of Probabilistic Forecasts of Munich Rents.}
		\label{fig:munich_dens}
	\end{figure}
	
	\subsubsection{Comparison to other approaches}
	
	To evaluate the prediction accuracy of \texttt{XGBoostLSS}, we compare the forecasts of the Munich rent example to the implementations available in $\it{gamlss}$, $\it{gamboostLSS}$, $\it{blackboostLSS}$\footnote{$\it{blackboostLSS}$ is a gradient boosting approach using conditional inference trees as base-learners.}, as well as to the Bayesian formulation of GAMLSS implemented in $\it{bamlss}$ by \citet{Umlauf.2017} and to Distributional Regression Forests of \citep{Schlosser.2018, Schlosser.2019} implemented in $\it{distforest}$. For all competing approaches, we use factor coding, instead of dummy-coding as for \texttt{XGBoostLSS}. We evaluate distributional forecasts in Table \ref{tab:comp} using the average Continuous Ranked Probability Scoring Rules (CRPS) and the average Logarithmic Score (LOG) implemented in the $\it{scoringRules}$ package of \citet{Jordan.2018}, where lower scores indicate a better forecast, along with additional error measures evaluating the mean-prediction accuracy of the models.\footnote{Scoring rules are functions $S(\hat{F},\veclatin{y})$ that assess the quality of forecasts by assigning a value to the event that observations from a hold-out sample $\veclatin{y}$ are observed under the predictive distribution $\hat{F}$, with estimated parameter vectors $\hat{\veclatin{\theta}} = (\hat{\theta}_{1},\ldots,\hat{\theta}_{K})^{\prime}$. See \citet{Gneiting.2007} for details.}
	
	\begin{table}[h!]
		\begin{center}
			\begin{threeparttable}
				\caption{Forecast Comparison}
				\begin{tabular}{rrrrrrr}
					\toprule
					Metric     & XGBoostLSS & gamboostLSS & GAMLSS & BAMLSS & DistForest & blackboostLSS \\  
					\midrule
					CRPS-SCORE & \textbf{1.1392} & 1.1541 & 1.1527 & 1.1509 & 1.1554 & 1.2315 \\  
					LOG-SCORE  & \textbf{2.1339} & 2.1920 & 2.1848 & 2.1656 & 2.1429 & 2.7904 \\ 
					MAPE       & \textbf{0.2450} & 0.2485 & 0.2478 & 0.2478 & 0.2532 & 0.2650 \\ 
					MSE 	   & \textbf{4.0687} & 4.1596 & 4.1636 & 4.1650 & 4.2570 & 4.5977 \\
					RMSE 	   & \textbf{2.0171} & 2.0395 & 2.0405 & 2.0408 & 2.0633 & 2.1442 \\ 
					MAE 	   & \textbf{1.6091} & 1.6276 & 1.6251 & 1.6258 & 1.6482 & 1.7148 \\ 
					MEDIAN-AE  & 1.4044          & 1.3636 & \textbf{1.3537} & 1.3542 & 1.3611 & 1.4737 \\ 
					RAE 	   & \textbf{0.7808} & 0.7898 & 0.7886 & 0.7890 & 0.7998 & 0.8322 \\ 
					RMSPE      & \textbf{0.3797} & 0.3900 & 0.3889 & 0.3889 & 0.3991 & 0.4230 \\ 
					RMSLE      & \textbf{0.2451} & 0.2492 & 0.2490 & 0.2490 & 0.2516 & 0.2611 \\ 
					RRSE       & \textbf{0.7762} & 0.7848 & 0.7852 & 0.7853 & 0.7939 & 0.8251 \\ 
					R$^{2}$    & \textbf{0.3975} & 0.3841 & 0.3835 & 0.3833 & 0.3697 & 0.3192 \\ 
					\bottomrule
				\end{tabular}
				\begin{tablenotes}
					\tiny
					\item \noindent Average Continuous Ranked Probability Scoring Rules (CRPS); Average Logarithmic Score (LOG); Mean Absolute Percentage Error (MAPE); Mean Square Error (MSE); Root Mean Square Error (RMSE); Mean Absolute Error (MAE); Median Absolute Error (MEDIAN-AE); Relative Absolute Error (RAE); Root Mean Square Percentage Error (RMSPE); Root Mean Squared Logarithmic Error (RMSLE); Root Relative Squared Error (RRSE); R-Squared/Coefficient of Determination (R$^{2}$). Best out-of-sample results are marked in bold (lower is better, except $R^{2}$).
				\end{tablenotes}
				\label{tab:comp}
			\end{threeparttable}
		\end{center}
	\end{table}

	\noindent All measures, except the Median Absolute Error, show that \texttt{XGBoostLSS} provides more accurate forecasts than the other implementations. The more accurate fit of our model compared to the other approaches might be attributed to the fact that \texttt{XGBoostLSS} automatically captures all potential interaction effects, while $\it{gamboostLSS}$, $\it{gamlss}$ and $\it{bamlss}$ are estimated as additive main effects models only that exclude interaction effects.\footnote{We haven't performed any parameter tuning for Distributional Regression Forests in our comparison, as the runtime for a forest with $T$ = 1,000 trees took around 3.5 hours on a Windows machine. $\it{gamboostLSS}$ is trained using parallelized 10-fold cross-validation to select the optimal number of iterations, with a run-time of around 9 hours.} Even though we could potentially include all interactions in these models, the number of effects that must be included can easily become unwieldy, especially for large $p$ data sets, as the structure of the data is typically unknown. The inherent tree structure of \texttt{XGBoostLSS} that automatically estimates all interactions provides therefore an advantage over existing models. To investigate the ability of \texttt{XGBoostLSS} of providing insights into the estimated effects on all distributional parameters, we compare its estimated effects to those of $\it{gamboostLSS}$.
	
	\vspace{-1em}
	
	\begin{figure}[h!]
		\centering
		\includegraphics[width=1.0\linewidth]{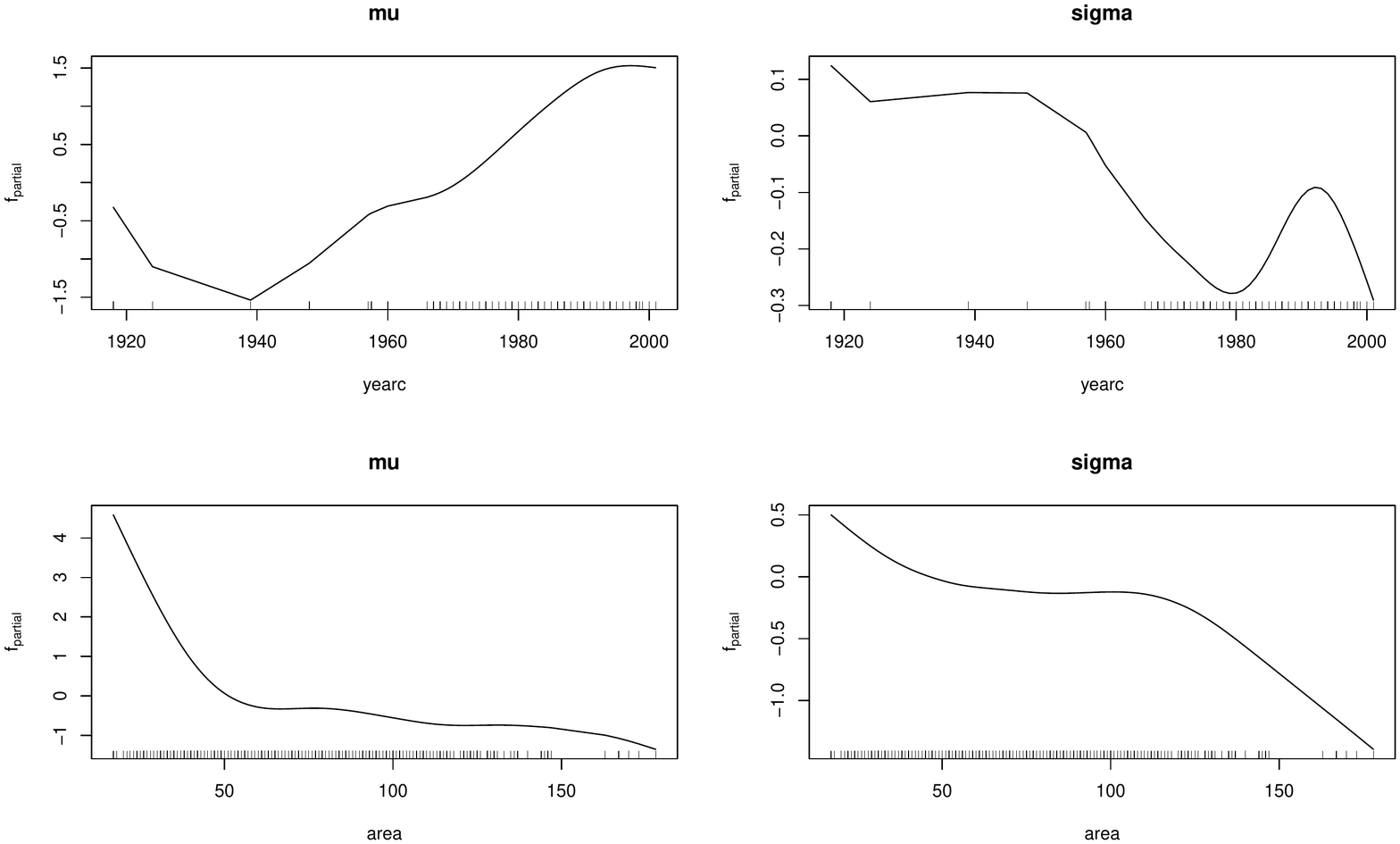}
		\caption{Estimated Partial Effects of $\it{gamboostLSS}$.}
		\label{fig:gamboostlss}
	\end{figure}
	
	\noindent Figure \ref{fig:gamboostlss} shows that all effects are similar to those of Figure \ref{fig:pdp} and therefore confirms the ability of \texttt{XGBoostLSS} to provide reliable insights into the data generating process.   
	
	\subsubsection{Expectile Regression}
	
	While GAMLSS require to specify a parametric distribution for the response, it may also be useful to completely drop this assumption and to use models that allow to describe parts of the distribution other than the mean. This may in particular be the case in situations where interest does not lie with identifying covariate effects on specific parameter of the response distribution, but rather on the relation of extreme observations on covariates in the tails of the distribution. This is feasible using Quantile and Expectile Regression. As with mean regression models, where the conditional mean is modelled as a function of covariates, both Quantile and Expectile Regression relate any specific quantile/expectile $\tau$ of the response to a set of covariates. Consequently, any desired point of the response distribution can be modelled, so that a dense grid of regressions yields a detailed description of the conditional distribution. Therefore, estimating and comparing parameter estimates across a different set of quantiles/expectiles allows for fully characterising the response distribution and for investigating the differential effect that covariates may have on different points of the conditional distribution. For our Munich rent analysis, Quantile/Expectile Regression yields additional insight compared to mean regression models, as they provide a richer description of the relationship between the rent of a flat and its attributing values for different values of $\tau$. In particular, standard models disregard important features of the data and yield an incomplete representation of the conditional distribution, as the conventional estimators are fixed for all quantiles/expectiles so that the estimated effects are averaged out over the response distribution. As such, Quantile/Expectile Regression is able to uncover heterogeneity across the conditional distribution, as the vector of regression coefficients $\veclatin{\beta}_{\tau}$ is allowed to vary with $\tau$, implying that latent factors nested in the regression coefficients are allowed to interact with the unobserved heterogeneity.
	
\begin{figure}[h!]
		\centering
		\includegraphics[width=1.0\linewidth]{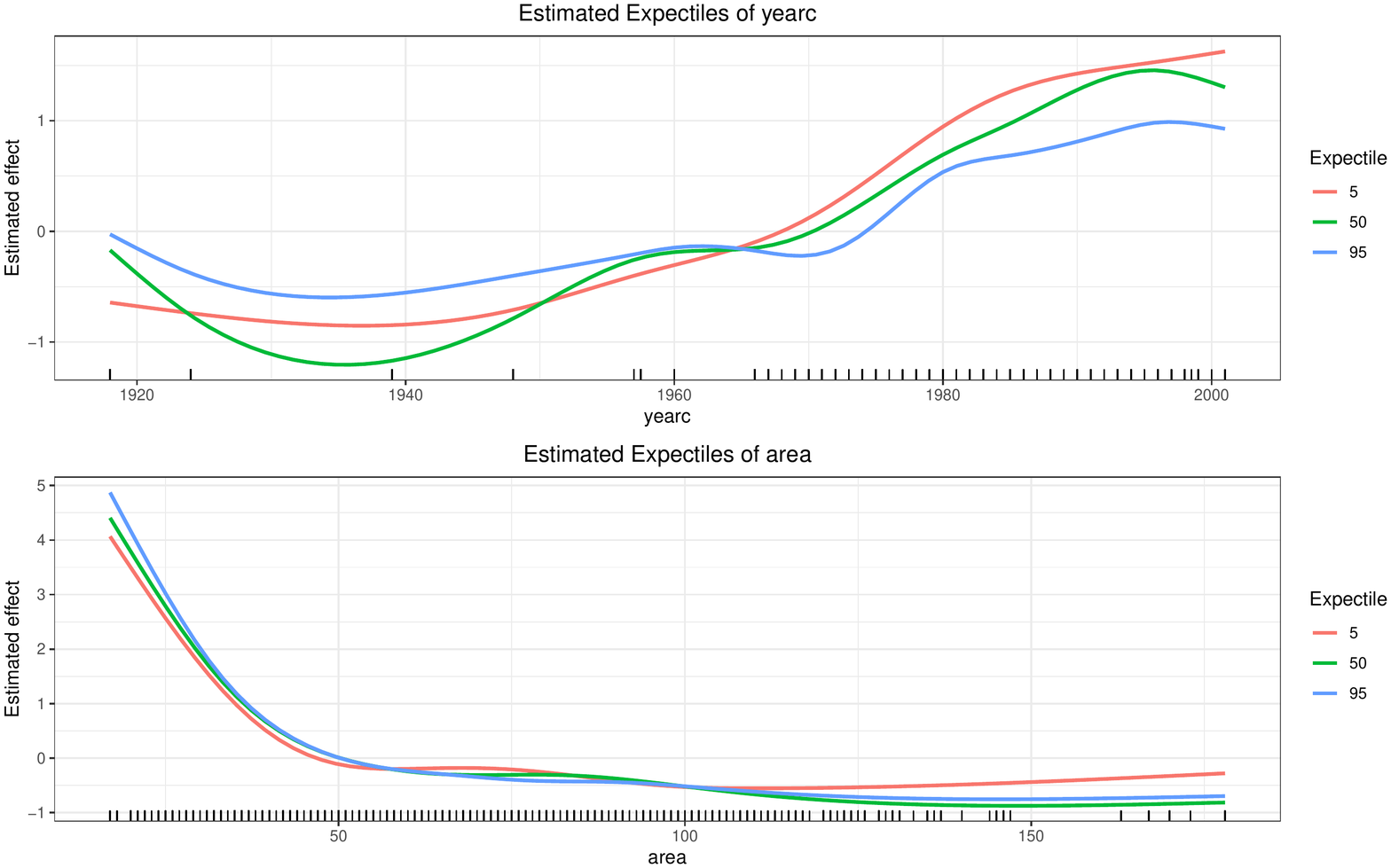}
		\caption{Estimated Partial Effects across different Expectiles.}
		\label{fig:expectile_pdp}
	\end{figure}

	As \texttt{XGBoostLSS} requires both the Gradient and Hessian to be non-zero, we illustrate the ability of \texttt{XGBoostLSS} to model and provide inference for different parts of the response distribution using Expectile Regression.\footnote{See \citet{Sobotka.2012} and \citet{Waltrup.2015} for further details on Expectile Regression.} As in the above examples, we use Bayesian Optimization to find the best hyper-parameter setting. Plotting the effects across different expectiles allows the estimated effects, as well as their strengths, to vary across the response distribution.\footnote{Even though excluded in theory, expectile crossing as shown in Figure \ref{fig:expectile_pdp} can occur, in particular with small data sets, as all expectiles are estimated separately. For suggestion on how to adjust the estimation process, we refer to \citet{Waltrup.2015} and the references therein.}
	
	\noindent Investigation of the feature importances across different Expectiles allows to infer the most important covariates for each point of the response distribution so that, e.g., effects that are more important for expensive rents can be compared to those from affordable rents.
	
	\newpage
	
	\begin{figure}[h!]
		\centering
		\includegraphics[width=1.0\linewidth]{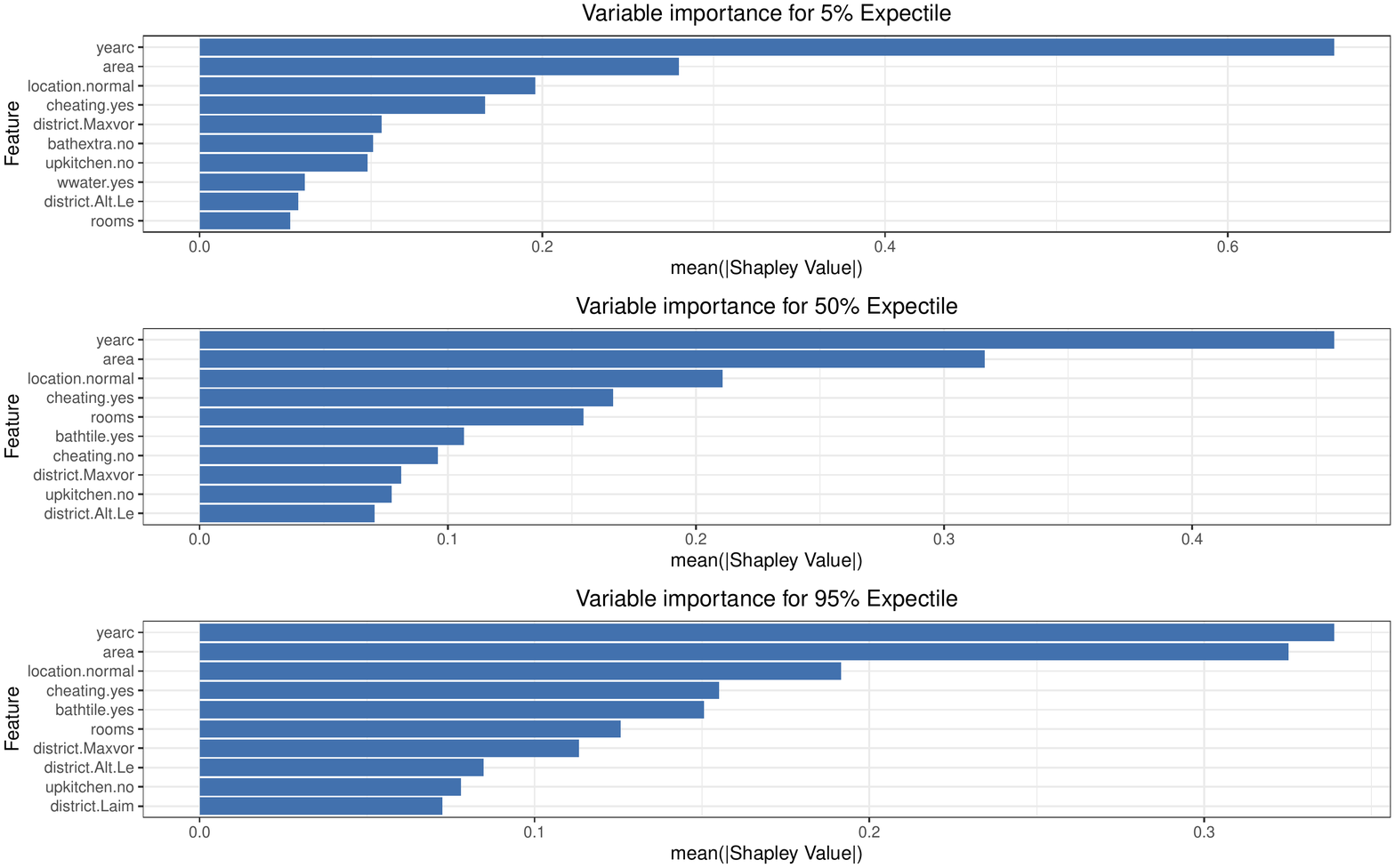}
		\caption{Mean Absolute Shapley Value across different Expectiles.}
		\label{fig:expectile}
	\end{figure}

	\section{Software implementation} \label{sec:implementation}
	
	In its current implementation, \texttt{XGBoostLSS} is available in \texttt{R} and made public soon following this link \faGithub\href{https://github.com/StatMixedML/XGBoostLSS}{StatMixedML/XGBoostLSS}. However, \texttt{XGBoostLSS} is generally compatible with all XGBoost implementations, i.e., \texttt{Julia}, \texttt{Python} and \texttt{Scala}. Extensions to \texttt{Julia} and \texttt{Python} are in progress.
	
	\section{Conclusion} \label{sec:conclusion}  
	
	\begin{quote} 
		\it{Assuming, as theory may tempt us to do, that covariates shift only the central tendency of the response, while variation around the central tendency remains unperturbed, is rarely plausible. Signal plus iid noise is a dangerous fiction}.\citep{Koenker.2013}
	\end{quote}
	
	The language of statistics is of probabilistic nature. Any model that falls short of providing quantification of the uncertainty attached to its outcome is likely to provide an incomplete and potentially misleading picture. While this is an irrevocable consensus in statistics, machine learning approaches usually lack proper ways of quantifying uncertainty. In fact, a possible distinction between the two modelling cultures can be attributed to the (non)-existence of uncertainty estimates that allow for, e.g., hypothesis testing or the construction of estimation/prediction intervals. However, quantification of uncertainty in general and probabilistic forecasting in particular doesn't just provide an average point forecast, but it rather equips the user with a range of outcomes and the probability of each of those occurring. In an effort of bringing both disciplines closer together, this paper extends XGBoost to a full probabilistic forecasting framework termed \texttt{XGBoostLSS}. By exploiting its Newton boosting nature and the close connection between empirical risk minimization and Maximum Likelihood estimation, our approach models and predicts the entire conditional distribution from which prediction intervals and quantiles of interest can be derived. As such, \texttt{XGBoostLSS} provides a comprehensive description of the response distribution, given a set of covariates. By means of a simulation study and real world examples, we have shown that models designed mainly for prediction can also be used to describe and explain the underlying data generating process of the response of interest. 
	
	We have seen that the GAMLSS framework provides the highest level of flexibility, both in terms of the variety of available distributions, as well as with respect to predictor specifications. However, with great power comes great responsibility. This is also true for distributional modelling, as its flexibility and complexity requires a careful investigation of the data set at hand, as well as the output generated. Even though this might be perceived as a drawback of the approach, we consider a careful analysis of the results and wrangling with the data as being at the heart of any sound analysis. Based on our current implementation, there are several directions for future research. Even though \texttt{XGBoostLSS} relaxes the assumption of observations being identically distributed, our model is not yet able to adequately incorporate dependencies between observations, e.g., time, longitudinal (where observations are nested in a hierarchical structure within groups or clusters) or space. Even though one could in principle add features that represent, e.g., the longitudinal structure or time, most machine learning models, however, are not directly applicable to non iid data without appropriate changes of the estimation process. One direct way to account for dependencies would be to replace cross validation with a dependency-respecting approach, such as time series or group cross-validation. However, this does not fully reflect all characteristics of the data in applications where there are dependencies between clusters, as is true for spatial data. A more promising approach would be to directly model the dependencies as part of the training of the model, as discussed in \citet{Hajjem.2011} or \citet{Sela.2012}. Another interesting extensions of distributional modelling was proposed by \citep{Klein.2015a, Klein.2016, Marra.2017} that extend the univariate case to a multiple response setting, with several responses of interest that are potentially interdependent. For high-dimensional settings, with a potentially large number of response variables, machine learning in general and decision trees/forests in particular can provide a viable alternative to existing approaches \citep{Segal.2011}.

% Bibliography (comment out for arvix) 
%\printbibliography

% Bibliography (comment out for working paper) 
\bibliography{references}

\end{document}